\documentclass[review]{elsarticle}

\usepackage{lineno,hyperref}
\usepackage{epsfig}
\usepackage{subfigure}
\usepackage{multirow}
\usepackage[table,xcdraw]{xcolor}
\usepackage{adjustbox}
\usepackage{hhline}
\usepackage{enumitem}
\usepackage{graphicx}
\usepackage{amsmath,amssymb} 
\usepackage{color}
\usepackage{booktabs, array}
 \usepackage[framemethod=tikz]{mdframed}
 \usepackage{lipsum}

\journal{Journal of \LaTeX\ Templates}

\begin{document}

\begin{frontmatter}

\title{Fine-grained Action Segmentation using the Semi-Supervised Action GAN}
%
%
%
%
\author[label1]{Harshala~Gammulle\corref{cor1}}
\ead{pranali.gammulle@hdr.qut.edu.au}

 \author[label1]{Simon~Denman}
\ead{s.denman@qut.edu.au}
 
\author[label1]{Sridha~Sridharan}
\ead{s.sridharan@qut.edu.au}
 
 \author[label1]{ Clinton~Fookes}
\ead{c.fookes@qut.edu.au}

 \cortext[cor1]{Corresponding author at: Image and Video Research Laboratory, SAIVT, Queensland University of Technology, Australia.}
 \address[label1]{Image and Video Research Laboratory, SAIVT, Queensland University of Technology, Australia.}

\begin{abstract}

In this paper we address the problem of continuous fine-grained action segmentation, in which multiple actions are present in an unsegmented video stream. The challenge for this task lies in the need to represent the hierarchical nature of the actions and to detect the transitions between actions, allowing us to localise the actions within the video effectively.
We propose a novel recurrent semi-supervised Generative Adversarial Network (GAN) model for continuous fine-grained human action segmentation. Temporal context information is captured via a novel Gated Context Extractor (GCE) module, composed of gated attention units, that directs the queued context information through the generator model, for enhanced action segmentation. The GAN is made to learn features in a semi-supervised manner, enabling the model to perform action classification jointly with the standard, unsupervised, GAN learning procedure. We perform extensive evaluations on different architectural variants to demonstrate the importance of the proposed network architecture, and show that it is capable of outperforming current state-of-the-art on three challenging datasets: 50 Salads, MERL Shopping and Georgia Tech Egocentric Activities dataset.  
  
\end{abstract}


\end{frontmatter}


\section{Introduction}

In the domain of human action recognition, continuous fine-grained action recognition is more challenging than utilising a pre-segmented video dataset where each video contains only a single action. The challenge for continuous fine-grained approaches stems from the need to represent the real-world hierarchical nature of the actions. For instance, within the broader category of actions related to cooking, the action `making salad' is composed of various related sub-actions such as `cutting vegetables', `add dressing', `mixing' etc. The main goal of fine-grained action segmentation is to predict what action is occurring at every frame in a video sequence. Hence, it requires an understanding of the different actions that can occur in similar settings. Often different actions can also involve the same objects (for eg: `Retract from shelf' and `Inspect product'), and also there are situations where the same action can be performed while interacting with different objects. Therefore, the problem is more challenging than recognising a single isolated action; the additional information such as semantic relationships among objects does not provide the same disambiguation in continuous fine -grained action recognition tasks as may be achieved in discrete action methods \cite{delaitre2011learning}. As a further complication, there are frames within the video that contain action transitions which do not belong to any action class, and are labelled as `background'  and should be identified as such.  

Most recent approaches for continuous action recognition are based on deep neural networks as they do not require feature engineering. However, manual effort is required to design an effective loss function. Furthermore, their performance is highly coupled with the database size and methods require very large labelled databases to generalise to different use cases. This causes a major hinderance for human action recognition task where labelled training data is scarce, particularly for many fine-grained human actions.

We are motivated by the recent advances in semi-supervised Generative Adversarial Networks (GAN) \cite{GAN2014}, where the model combines a supervised classification objective with an unsupervised GAN objective. In another line of work by Ahsan et. al\cite{GAN_action_2} have suggested a semi-supervised learning strategy, where they do not directly couple the adversarial GAN loss together with the action classification loss and do not perform joint training. Instead, they first train the model using adversarial GAN loss and then fine-tune it for action classification. Hence, the task specific loss function learning paradigm of the GAN is not aware of the end goal of the system, i.e. action classification.

In this paper, we present a novel semi-supervised generative adversarial network architecture for continuous fine-grained action segmentation. The generator's role is to learn an intermediate representation, an `action code', that has high discriminative power to aid the classification task performed by the discriminator. Using this mapping we simplify the action recognition process, and as the action codes are unique to a specific action class, the generator directly contributes to the action recognition and the task of the classifier is simplified.

\begin{figure}[!ht]
        \centering
        	\includegraphics[width=0.9\textwidth]{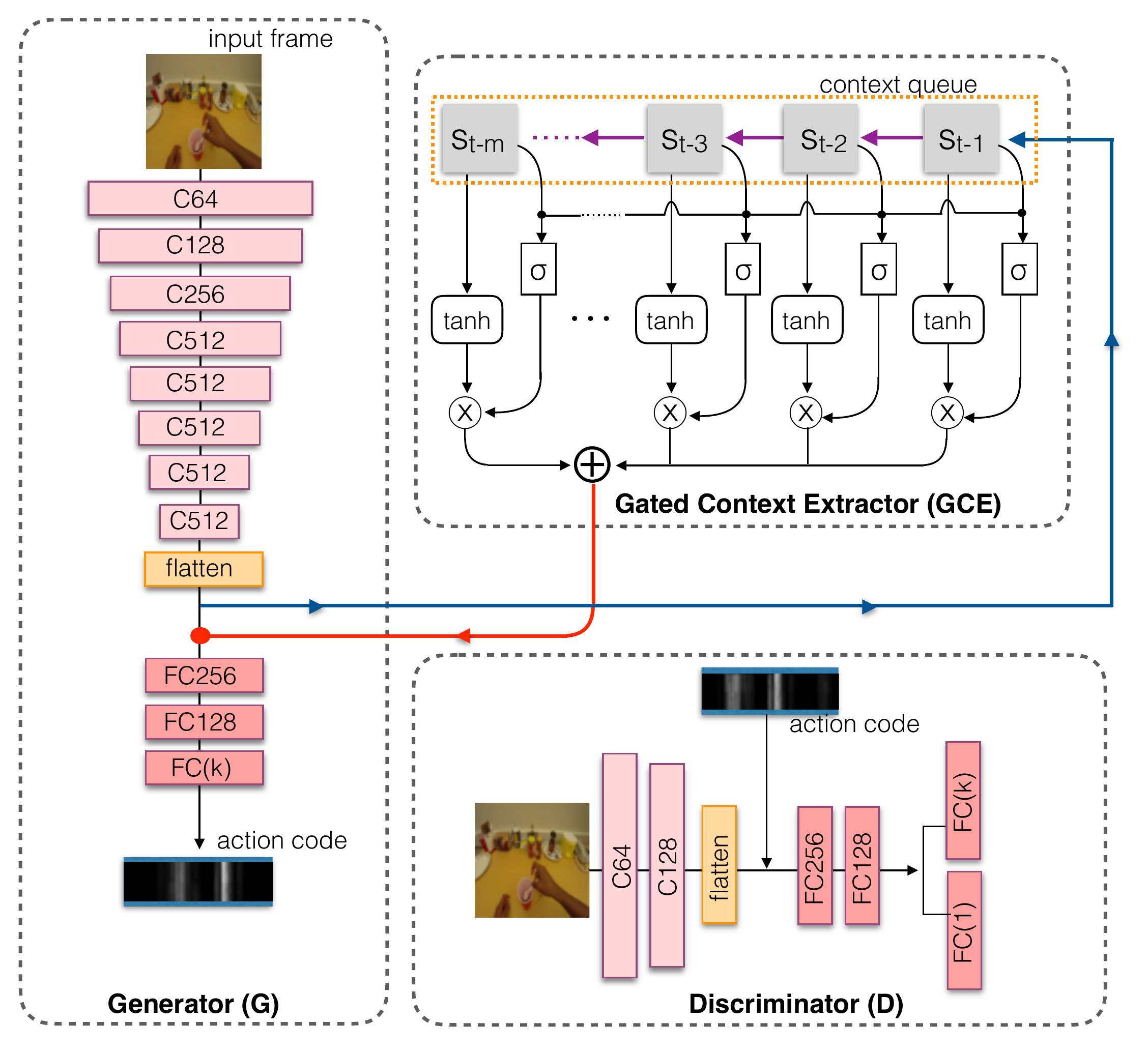}
	\vspace{-4mm}
	\caption{The proposed \textit{SSA-GAN}: G is trained to generate action codes (action representation) of each input frame while D performs real/fake validation and classification on the frame and action code. The GCE module maintains a context queue (contains hidden layer features from the previous m frames), which is passed through a gated network that connects to the generator to enhance the action generation process. CX are convolutional\_BatchNorm\_ReLu layer groups with X filters, \cite{Isola_CVPR2017}, and FC(k) and FC(1) denote fully connected layers with softmax activations of size 1 and k respectively.}
	\label{fig:system_diagram}
	\vspace{-3mm}
\end{figure}

Figure \ref{fig:system_diagram} illustrates the proposed action segmentation framework. Due to the hierarchical structure of human actions, there exists a high probability of certain sub-action classes co-occurring. To capture such complex temporal relationships, the generator of the proposed model is coupled with the gated context extractor (GCE) module, which maintains a queue called `context queue' containing previous frame features. Therefore, the action code generation process is also influenced by the signal that passes through the GCE. Through this process the overall model becomes recurrent where the decision making process at the present time step leverages embedded information from the previous frames.

Our GCE model is inspired by the work in \cite{gatedfusion}, where the authors propose gated operations for multimodal information fusion. However, to the best of our knowledge, no previous works have considered using a queue like structure for temporal modelling. Furthermore, in \cite{gatedfusion} the authors only consider the gated operations when there are two elements. We extend these operations to a larger scale and train the model to utilise historical information from the distant past. 
 
The main contributions of the proposed work can be summarised as follows:

\begin{itemize}
\item We introduce a novel recurrent semi-supervised action GAN (SSA-GAN) model for continuous fine-grained human action segmentation. 
\item We propose the incorporation of a novel Gated Context Extractor (GCE) to model long term temporal relationships among DCNN features stored in the context queue, capturing high level contextual information. 
\item We show the proposed model outperforms the current state-of-the-art methods for three challenging public datasets.
\end{itemize}

The rest of the paper is organised as follows. In Section \ref{sec:lit_rev} we discuss related work on action segmentation with deep architectures and Section \ref{sec:methods} contains a description of the methods used in developing the proposed network architecture. In Section \ref{sec:experiments} we provide details on network architecture and training with a performance evaluation on three challenging datasets and an ablation study to show the importance of different components of the the proposed \textit{SSA-GAN} model. A discussion analysing performance gain of the proposed method is presented in Section \ref{sec:discussion} and Section \ref{sec:conclusion} concludes the paper.

\section{Related Work}
\label{sec:lit_rev}

Human action recognition approaches can be categorised into two types: methods that are discrete and operate on either image \cite{imageBased1} or pre-segmented videos \cite{twoStream,gammulle2018multi,PR_discActions3}; and methods that operate over continuous fine-grained action videos \cite{merlshopping,gammulle2019coupled,PR18_fine_grained2}. Even though discrete methods have demonstrated greater performance \cite{PR_discActions2, PR17_discActions1}, they are disconnected from real world scenarios that are always composed of fine-grained actions. This has been the motivation for researchers to focus on methods that process continuous fine-grained videos. These approaches are more relevant to real world applications such as detecting threats from a surveillance video, as they are able to keep track of previously observed actions and exploit the relationships between consecutive actions.

Methods have been implemented either as hand crafted feature based \cite{wang2011} or deep network based \cite{twoStream,PR_discActions4, PR_discActions1} approaches. Recent works have preferred deep network based models as they do not require feature engineering. Among these approaches for fine-grained action segmentation, Ma et al. \cite{PR18_fine_grained}, introduced a method based on Convolutional Neural Networks (CNN), which takes the human pose and human part position based image patch sequences and utilises vectors of locally aggregated descriptors (VLAD) to encode the final pooling layer output. Rich action information is provided by the variety of different scales of appearance and motion patch data that is processed in a CNN model. Furthermore, the addition of the VLAD encoding mechanism facilitates more effective action description and improves action segmentation. Another segmental model, termed a Temporal Convolutional Network (TCN), is introduced by Lea et al. \cite{leaCVPR} and is capable of capturing segmentation features such as action durations, pairwise transitions between segments and long range action dependancies while performing human action segmentation. Their encoder-decoder network (ED-TCN) hierarchically models action instances with the use of temporal convolutions, pooling, and upsampling operations while the dilated TCN model uses dilated convolutions with added skip connections between layers. This idea is further extended by Lei et al. \cite{TDRN2018} who replace the temporal convolution layers of the ED-TCN model with deformable temporal convolutions, allowing the model to capture fine-scale temporal details. In \cite{lea2016}, the proposed spatio-temporal CNN (ST-CNN) architecture is capable of capturing information such as object states, their relationships and their changes over time. In \cite{tricornet}, the authors introduced an encoder-decoder architecture which is a hybrid of temporal convolutional and recurrent models. However all above mentioned deep networks require manual human effort to design effective losses. Utilising Generative Adversarial Networks (GAN) offers a means to overcome this as they are capable of learning a loss function, and have other benefits such as no inference being required during learning, and a wide variety of factors and interactions can easily be incorporated into the model \cite{condGAN}. 

The GAN was originally introduced by Goodfellow et al. \cite{GAN2014} and has increased in popularity in various research areas over the last few years. GAN based models are capable of learning an output that is difficult to discriminate from real examples, and learn a mapping from input to output while learning a loss function to train the mapping. Hence, the GAN learning process is formulated as a min max game between the generator (G) and the discriminator (D). In the GAN learning framework we do not define a loss for the synthesised examples, rather this is learnt automatically via the ability of the synthesised examples to fool D. Hence, the frameworks leans a task specific loss. Given this ability, GANs have been applied for diverse computer vision problems such as state prediction \cite{zhou2016}, future frame prediction \cite{mathieu2015}, product photo generation \cite{yoo2016}, and inpainting \cite{inpainting}. Most of these image-to-image generation based models are based on an extension to the GAN, namely the conditional GAN (cGAN) \cite{condGAN}, where both the generator and the discriminator are conditioned with extra information such as class labels or data from other modalities. However, it is not always possible to obtain large numbers of labelled samples to train these models. In such cases semi-supervised GANs \cite{semi_supGAN_3, denton2016semi} are convenient as they are able to learn models from both labelled and unlabelled data. 

A limited number of methods have been introduced using GANs for the discrete human action recognition task \cite{GAN_action_1, GAN_action_2}. The method of \cite{GAN_action_1} utilised GANs only for generating masks to detect the actors in the given input frame. Then the relavant action classification is performed via a CNN. This method is prone to difficulties with the loss function as noted previously. 

The method of \cite{GAN_action_2} proposes a Semi-Supervised GAN architecture for discrete action recognition. The generator is given sample frames of the video from which it learns spatial features of action categories and the discriminator learns to classify the relevant action classes of the input frames. However, as yet, no methods have been introduced that use GANs for continuous fine-grained action recognition. When modelling human actions in continuous videos we have to consider both temporal and visual features to capture long-term relationships. 

In a different but related line of work, tracking temporal consistency has been formulated as minimising the divergence between consecutive frames. Specifically, in \cite{zhang2018robust} the authors utilise a multi-task learning platform to make the learned objectives from neighbouring frames closer to each other, while in \cite{fernando2018tracking} the authors make the objective follow the anticipated future human behaviour. In contrast, while following the same fundamentals of temporal consistency, we propose a method to capture it via queuing historic frames and utilising a gated attention framework to extract the salient temporal information that should be considered when making decisions regarding the current frame. 

\section{Methodology}
\label{sec:methods}

GANs are generative models that are able to learn a mapping from a random noise vector $z$ to an output vector $y: G:z\rightarrow y$ \cite{GAN2014}. In our proposed method, we utilise the conditional GAN \cite{condGAN}, an extension of the GAN that has the ability to learn a mapping from the observed image $x_{t}$ at time $t$ and a random noise vector $z_{t}$ to $y_{t}: G:\{x_{t},z_{t}\}\rightarrow y_{t}$ \cite{Isola_CVPR2017}, where $y_{t}$ is the generator output at time step t. GANs are principally composed of two components: the Generator (G) and the Discriminator (D), which compete in a two player game. G tries to generate data that is indistinguishable from real data while D tries to distinguish between real and generated (fake) data. Therefore, the ultimate target of the model G is to fool the model D.     

We introduce a conditional GAN based model, Semi-supervised Action GAN (\textit{SSA-GAN}), for continuous fine-grained action segmentation. The proposed \textit{SSA-GAN} model couples spatial and temporal information and through the semi-supervised architecture it is able to perform action classification via the discriminator. Here, unlike a typical GAN that utilises only spatial information, the generator of our proposed model gains information in the form of an action code, which is an intermediate representation learned by the network. We use this approach as areas such as action segmentation involve long video sequences, and long-term feature relationships between frames are crucial. 

In Section \ref{ac_code}, we describe the action code format that the GAN is trained to generate and the importance of such intermediate action representations for the prediction task; Section \ref{semi} describes the semi-supervised GAN architecture and how it is capable of performing direct classification through the GAN model; in Section \ref{obj} we explain the objectives behind our models and in Section \ref{context_eq} we explain the Gated Context Extractor (GCE) model that captures the long-term temporal relations and utilises them within the overall process. 

\subsection{Action codes}
\label{ac_code}

\begin{figure}[!ht]
    \centering
    \subfigure[] {\includegraphics[width=.42\linewidth]{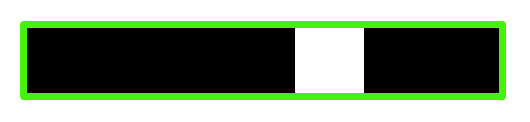}}
     \subfigure[] {\includegraphics[width=.43\linewidth]{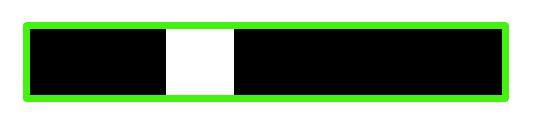}}
     \vspace{-4mm}
    \caption{Sample ground truth action codes with $k=7$ (i.e. 7 actions), black regions represent the value 0 while white regions represent the value 255. For the code in (a), $y=5$ and for the code shown in (b) $y=3$. A green border is shown around the codes for clarity, this is not part of the code and is only included to aid display. Codes are of size 1 x k pixels.}
    \label{fig:action_codes}
\end{figure}

The aim of our generator model is to synthesise an intermediate action representation, called an `action code', to represent the current action in each input frame. The generator maps dense pixel information to this action code. Hence having a one hot vector is not optimal. Therefore we scale it to a range from 0 to 255 giving more freedom for the action generator and discriminator to represent each action code as a dense vector representation, 
\begin{equation}
y_{t}  \epsilon  {\rm I\!R}^{1 \times{k}},
\label{eq:1}
\end{equation} 

where k is the number of action classes in the dataset. Several works \cite{bora2017compressed,infogail_nips} have shown the importance of using such representations with GAN architectures. In our work the action code is influenced by both the adversarial loss and the classification loss. We give more attention to the classification process. Hence the action codes must be informative for classification. In Figure \ref{fig:action_codes} we show some examples for the ground truth action codes for a scenario where there are 7 action classes. It is essential to state that this idea of the proposed action codes have been used as an example for a simple embedding of the action representation. However any distinct representation can be utilised here as long as they are unique for each action class. See Section \ref{sec:discussion} for more details.  

\subsection{Semi-Supervised GAN architecture}
\label{semi}

Our \textit{SSA-GAN} network is composed of a conditional GAN architecture that enables semi supervised learning. Semi-supervised learning has been added to the network by combining a supervised objective and an unsupervised objective during training \cite{denton2016semi}. The generator of a standard GAN is utilised to generate outputs that are more similar to the ground truth data while the aim of the discriminator is to distinguish between the ground truth (real) and generated (fake) data. Even though the ground truth labelled data is provided to the discriminator, the generator is not fed with any labelled data. It learns the output through the overall loss at each time step. Hence, for a standard GAN model only the real/fake labels are provided. However, classifying real/fake is not the main goal of the model. The main goal is the generation of optimal outputs that are similar to the real data. Hence the learning procedure of a standard GAN, and in particular the generator, is unsupervised \cite{Isola_CVPR2017} in that it does not utilise labelled data during the training procedure. In the proposed work we utilise an additional classification model coupled with the discriminator. When considered individually, classification is a supervised task where the learning is purely based on labelled data. Therefore, with the addition of a supervised model to the unsupervised GAN, the overall architecture becomes semi-supervised \cite{denton2016semi}. This has been achieved by enabling the discriminator to perform classification on action class labels that are available in the datasets. Therefore, in addition to learning the real/fake examples, the discriminator also learns the probabilities of each of the original dataset classes that it has been trained on; and the unlabelled data used for real/fake verification is able to support learning the hierarchical nature of the input. The real/fake verification component also enables the discriminator to send a signal back to the generator to improve its ability to generate realistic action codes. 

\subsection{Objectives}
\label{obj}

 Conditional GANs are capable of learning a mapping from input to output while learning a loss function to train this mapping. Therefore they are useful for problems that require varying loss formulations. The objective for the conditional GAN can be defined as, 

\begin{equation}
\begin{split}
L_{cGAN}(G,D)= \mathop{min}_{G} \mathop{max}_{D} \hspace{1mm} \mathbf{E}_{x_t,y_t\sim p_{data}(x,y)}[\log D(x_t,y_t)]+ \\ \mathbf{E}_{x_t\sim p_{data}(x),z_t\sim p_{z}(Z)}[\log(1-D(x_t,G(x_t,z_t))],
\end{split}
\label{eq:2}
\end{equation}   
where $G(x_{t},z_{t})$ refers to the generator output given input image $x_t$ and the noise distribution $z_t$ at time instance t.
These networks have been mainly used as a general purpose solution for image-to-image translation problems \cite{Isola_CVPR2017}, and as suchvrequires adaptations for use in classification. The use of a semi-supervised conditional GAN architecture is preferable as the discriminator is capable of learning the label classification as well as learning to verify real/fake data. 
Let the labelled dataset be ${(x_{1},k_{1}), (x_{2},k_{2}),...,(x_{n},k_{n})}$ where $k_t$ is the label of input image $x_t$ and $D_{c}(x)$ is the output of the classifier head in the discriminator. Then the objective function of the semi-supervised model can be defined as follows,

\begin{equation}
\begin{split}
L_{cGAN}(G,D)= \mathop{min}_{G} \mathop{max}_{D} \hspace{1mm}\mathbf{E}_{x_t,y_t\sim p_{data}(x,y)}[\log D(x_t,y_t)]+ \\ \mathbf{E}_{x_t\sim p_{data}(x),z_t\sim p_{z}(Z)}[\log(1-D(x_t,G(x_t,z_t))]+ \lambda_{c} \mathbf{E}_{x_t,k_t\sim p_{data}(x,k)}[\log D_{c}(k_t|x_t)]  
\end{split}
\label{eq:3}
\end{equation}   

Here the balance between the classification loss and the adversarial loss is achieved through a hyper parameter, $\lambda_{c}$.

Fine-grained action segmentation becomes challenging as these datasets are usually composed of visually similar actions that belong to different action classes. There is a higher chance of actions being visually similar when they appear consecutively in the video sequence. In particular, the frames constituting the `background' are highly visually similar to the surrounding actions. In such scenarios the use of additional context information becomes supportive for the learning task. Many visual recognition approaches critically rely on context \cite{context,Shapovalova2011,santofimia2014,vrigkas2015}. Therefore, we utilise context information provided by the previous action frames which will be discussed in Section \ref{context_eq}. This context information is stored and handled by the gated context extractor (GCE), $\pi$.

After coupling the generator with the context extractor, the objective function can be defined as,

\begin{equation}
\begin{split}
V_{G_{A},D}=\mathbf{E}_{x_{t},y_{t}\sim p_{data}(x)}[\log(D(x_{t},y_{t}))]+ \\ \mathbf{E}_{x_{t}\sim p_{data}(x),z_{t}\sim p_{z}(Z),c_{t}\sim p_{\pi}}[\log(1-D(x_{t}, G(x_{t},z_{t},c_{t})))] + \\ \lambda_{c} \mathbf{E}_{x,k\sim p_{data}(x,k)}[\log D_{c}(k|x)]  ,
\end{split}
\label{eq:4}
\end{equation} 
where $G(x_{t},z_{t},c_{t})$ refers to the generator output given input image $x_t$, noise distribution $z_t$, and context vector $c_t$ produced by the GCE at time instance t.

\subsection{Gated Context Extractor (GCE)}
\label{context_eq}

We observe that fine-grained action videos are generally composed of related actions. Hence, information from previous frames is beneficial when predicting the current action. When performing a specific activity such as `preparing salad' in the 50 Salads dataset, the chance of the occurrence of some actions (e.g. `place lettuce into bowl') after a particular action (`cut lettuce') can be higher. For this reason, in order to capture long-term temporal dependancies, the GCE module is allowed to store information extracted from the generator for the previous $m$ frames, which are fed into the generator through a series of Gated Attention Units (GAU), inspired by \cite{gatedfusion}. The proposed GAU can be expressed as follows,

\begin{equation}
h_{t-j}= tanh(W^h_{t-j} s_{t-j}),
\label{eq:5}
\end{equation}  
where $s_{t-j}$ is the hidden state representation and the weight $W^h_{t-j}$ is learnt jointly with the other components. 
 First we encode the stored hidden state representation $s_{t-j}$ by passing it through a $tanh$ function. Then a sigmoid function, $\sigma$, is used to determine the information flow from the present state $s_{t-j}$ by attending over all the stored information,  
    
\begin{equation}
q_{t-j}=\sigma(W^q_{t-j} [s_{t-m}, ... ,s_{t-1}]),
\label{eq:6}
\end{equation}  
 
  where [.;.] denotes concatenation. Each of these units act as a gate function which controls the amount of information transferred to the final output of the GCE. We achieve improved temporal modelling compared to LSTM \cite{LSTM1997} cells where the output depends only on the immediately preceding cell state. In contrast to an LSTM,  we consider the entire stored sequence when determining the output of each gate. Then we multiply the embedded state from Equation \ref{eq:5} with the output of the gate such that,

\begin{equation}
r_{t-j}=h_{t-j}\times{q_{t-j}}, 
\label{eq:7}
\end{equation}  

and determine the final output, $c_{t}$, of the GCE module by aggregating all the outputs of individual gates as, 

\begin{equation}
c_{t}=\sum_{j=1}^{m} r_{t-j}.
\label{eq:8}
\end{equation}

\section{Experiments}
\label{sec:experiments}

\subsection{Datasets}
\label{data}

We evaluate our proposed \textit{SSA-GAN} model on three challenging fine-grained action datasets similar to \cite{leaCVPR}, containing up to 17,310 frames per video. Hence, when evaluating, it requires us to consider all the frames within each video sequence, understanding the actions together with the action transitions.  

\textbf{The University of Dundee 50 Salads Dataset \cite{50salads}} is composed of 50 video sequences of 25 subjects, where each subject prepares two salads in two different videos. Videos are captured by a static RGBD camera pointed at the subject, with a duration of 5 to 10 minutes. Multi-modal data including depth and accelerometer data is provided alongside time synchronised videos, although we only use the video data. The 50 Salads contains videos of higher level action classes that are formulated by a combination of multiple fine-grained actions. For example, the higher level class `cut\_and\_mix\_ingredients' is composed of multiple fine-grained actions (also termed mid-level action classes) such as `peel\_cucumber', `cut\_cucumber', `place\_cucumber\_into\_bowl' etc. Following the work in \cite{leaCVPR}, we utilise 17 available mid level action classes for action segmentation
When training the model all 17 action classes are used with the background class frames. In each video sequence subjects perform around 12 different actions to prepare a salad.    

\textbf{ The MERL Shopping Dataset \cite{merlshopping}} contains 96 videos of 32 subjects shopping from grocery-store shelving units. Each subject performs in three different videos of two minutes duration, which are obtained via a static overhead HD camera. The dataset contains a total of five action classes and the background class. 
Each video contains different combinations of the 6 classes through out the video.

\textbf{The Georgia Tech Egocentric Activities Dataset \cite{GTEA_1}} contains videos of four subjects performing seven different daily activities: preparing a hot dog sandwich, instant coffee, peanut butter sandwich, jam and peanut butter sandwich, sweet tea, coffee and honey, and a cheese sandwich. These videos are recorded from a head mounted GoPro camera which is fixed to a baseball cap worn by the subjects. The total number of frames in the dataset is 31,222 and all frames have been utilised to evaluate our proposed model. The dynamic egocentric camera setting of this dataset is significantly different to static top view of the previous 2 datasets. We utilise 11 action classes defined in \cite{gtea_evalpaper} including the background class.         

\subsection{Metrics}
\label{met}

The evaluation of the proposed model uses both segmentation and frame wise accuracy metrics. Frame wise metrics are widely used in many works \cite{lea2016seg,leaCVPR,50salads}, however, as explained in \cite{leaCVPR}, models that gain similar frame wise accuracies still can show large variations when visualising their performance due to different segmentation behaviour. Hence, to fully describe the action segmentation performance of the proposed model we also utilise segmentation metrics such as mean average precision with midpoint hit criterion (mAP@mid) \cite{merlshopping}, Segmental F1 score (F1@k) \cite{leaCVPR} and segmental edit score (edit) \cite{lea2016}.   

\subsection{Network Architecture and Training}
\label{opt}

\begin{figure}[!ht]
\begin{center}
    \subfigure[Queue size $m$ against the accuracy]{\includegraphics[width=.412\textwidth]{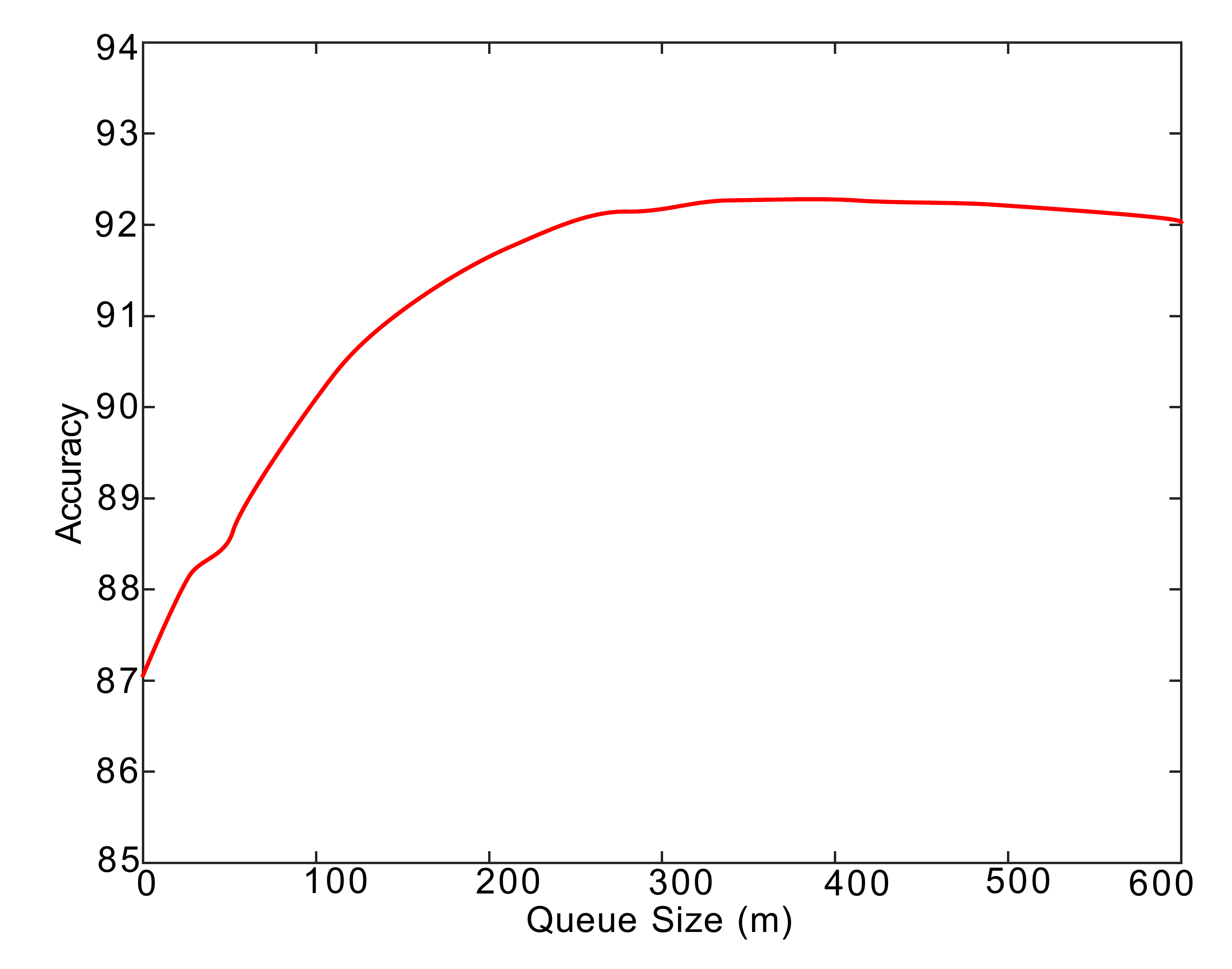}} %
     \subfigure[$\lambda$ against the accuracy]{\includegraphics[width=.4\textwidth]{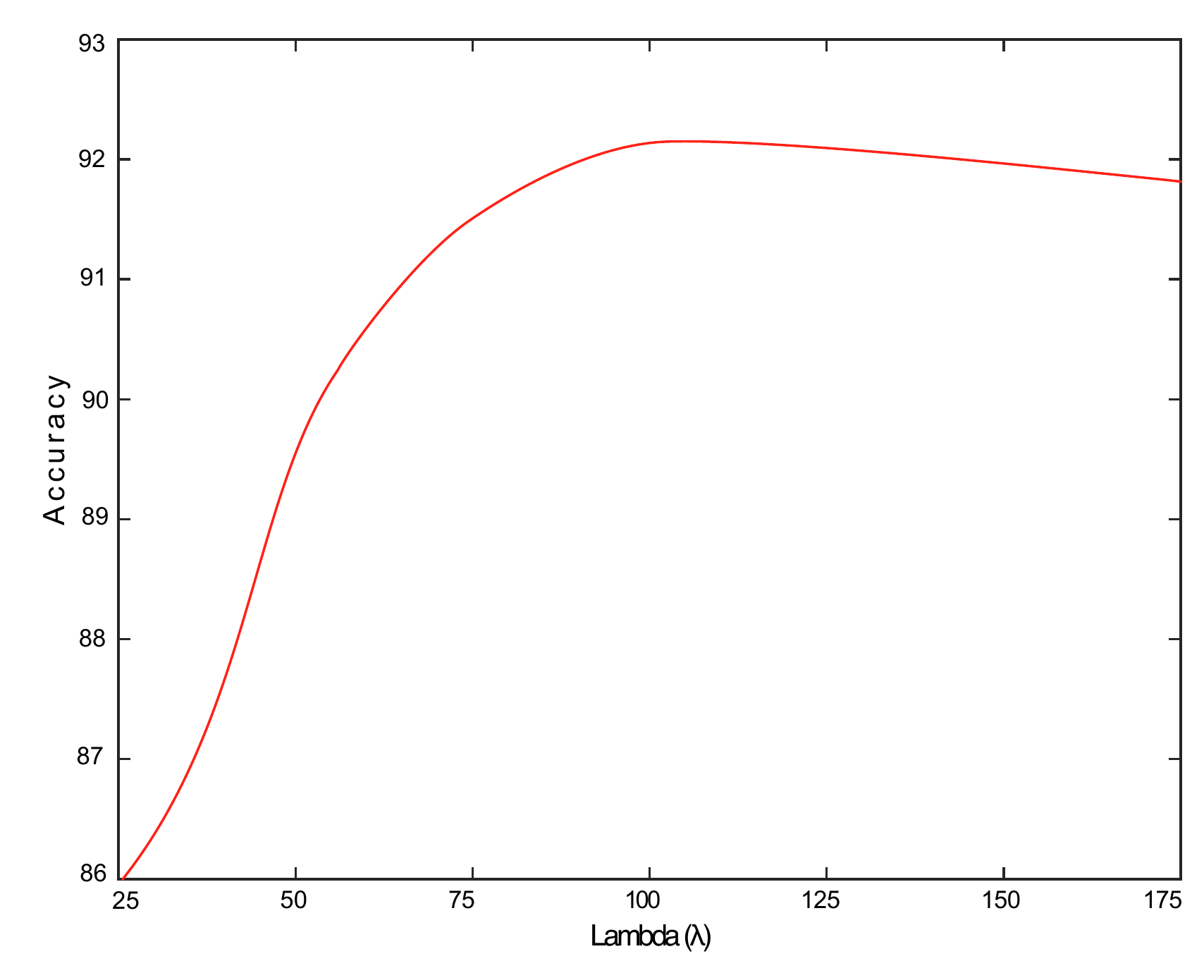}} %
 \end{center}
 \vspace{-4mm}
  \caption{Hyper-parameter evaluation. We evaluated the distribution of accuracy for different queue size $m$ and $\lambda$ values for the validation set of MERL Shopping dataset}
  \label{fig:hyper_parameter}
\end{figure}

The network architecture is defined in Figure \ref{fig:system_diagram}. We evaluated different queue sizes, $m$, and the optimal size of $m=400$ is determined experimentally. In Fig. \ref{fig:hyper_parameter} (a) we show the accuracy against different queue sizes for the test set of the MERL Shopping dataset and we set $m=400$ as it offers the best accuracy. We follow the training procedure of \cite{Isola_CVPR2017}, alternating between one gradient decent pass for the discriminator and one for the generators using mini batch SGD (32 examples per mini batch) and the Adam optimiser \cite{adam2015}, with an initial learning rate of 0.1 for 250 epochs, and 0.01 for the next 750 epochs. For the discriminator model, we take (batch\_size)/2 generated (fake) action codes and (batch\_size)/2 ground truth (real) action codes, where ground truth codes are created manually. We utilise Keras \cite{keras} with Theano \cite{theano} as the backend to implement our proposed model.  

Similarly, we evaluated our model with different values of $\lambda$ between 25 and 175. The model gains it's highest frame-wise accuracy for the MERL shopping dataset at $\lambda=100$. This value of $\lambda=100$ is used when evaluating all three datasets. We do not evaluate on $\lambda=0$ as it completely eliminates the classification objective from the overall objective function. When the value of $\lambda$ decreases from 100 to 25, the accuracy drops from 92.1 to 86 showing that the classification objective has more of an effect on the model performance than the GAN objective. When $\lambda$ is a very large value ($>$100), the model focuses more on the classification task. Therefore, the effect of the GAN objective will be considerably low. According to Figure \ref{fig:hyper_parameter} (b), the frame-wise accuracy tends to decrease when $\lambda$ is larger than 100. From this, it is evident that both the GAN and the classification objectives contributed to the gain in accuracy. 

\subsection{Results}

Table \ref{tab:tab_new} presents results for the proposed approach along with the state-of-the-art baselines. For all three datasets we consider the models proposed in \cite{leaCVPR} as baselines. In \cite{leaCVPR}, the authors introduce two networks, namely encoder-decoder TCN (ED-TCN) and the dilated TCN, where the ED-TCN utilise pooling and up sampling to capture long range temporal patterns while the dilated TCN utilises dilated convolutions.

For the 50 salads and Georgia Tech Egocentric activity datasets we also compare to the results  obtained by Lea et al. in \cite{lea2016} for their spatial CNN and spatio-temporal CNN (ST-CNN). They propose a CNN architecture capable of capturing information such as object states, their relationships and their changes over time. We also compared the results to the Bi-LSTM model of \cite{leaCVPR}, TricorNet model proposed in \cite{tricornet} and the TDRN \cite{TDRN2018} model. The TricorNet model is a hybrid of temporal convolutional and recurrent models containing an encoder-decoder architecture. The TDRN model could be seen as an extension of ED-TCN where the authors replace the temporal convolution layers of the ED-TCN model using deformable temporal convolutions, allowing the model to capture fine-scale temporal details, in contrast to the fixed temporal receptive size of ED-TCN. For the MERL shopping dataset we compare the proposed approach against the `MSN Det' and `MSN Seg' methods introduced by Singh et al. \cite{merlshopping}. 

Furthermore, to better demonstrate the strengths of the automatic feature learning ability attained by these deep learning (but non-GAN) methods we compare these models as well as the proposed Semi-Supervised Action GAN (\textit{SSA-GAN}) model with traditional non-deep learning models that utilise hand crafted features. For comparisons on the 50 salads dataset we use two models that use Improved Dense Trajectories (IDT) \cite{wang2013action} together with a Language Model (LM) \cite{IDT_LM} and Conditional Random Field (CRF) \cite{rupprecht2016sensor} to segment the temporal actions. For the Georgia Tech Egocentric activity dataset, as non-deep learning based models we utilise two SVM classifiers trained on the well known Space-Time Interest Points (STIP) \cite{laptev2005space} and Scale-Invariant Feature Transform (SIFT) \cite{lowe2004distinctive} features. We also use a model utilising hand motion, hand location, hand pose and foreground object details as features, which recognises the actions using a SVM \cite{gtea_evalpaper}. Due to the unavailability of non-deep learning based baseline model results for the MERL Shopping dataset, we were unable to perform such a comparison for this dataset. 

When considering the results presented in Table \ref{tab:tab_new}, we observe that none of the hand crafted feature based approaches (non-deep learning) have been able to attain results comparable with either the deep learning (non-GAN) methods, or the proposed SSA-GAN approach, convincingly demonstrating the importance of the automatic feature learning process. 

Among the deep learned models we observe better performance from ED-TCN, Bi-LSTM, TricorNet and TDRN compared to other baselines owing to their improved temporal modelling. We observe similar frame wise accuracies for the Spatial CNN, dilated TCN, ST-CNN, Bi-LSTM, ED-TCN, TricorNet and TDRN models. However, we see significant variations between F1-scores, mainly due to over segmentation. 

The proposed \textit{SSA-GAN} achieves better performance compared to the baselines in all considered metrics for all datasets. We observe a 15.2\% and 11.5\% increase in frame wise accuracy compared to TricorNet for the 50 Salads and Georgia Tech Egocentric datasets respectively. For the MERL Shopping dataset the frame wise accuracy is increased by 13.1\%. We observe similar performance for other metrics.

The proposed semi supervised GAN framework is capable of learning the hierarchical structure of the input frames along with the generated action codes, enabling improved classification of the action classes. Furthermore, in contrast to the Bi-LSTM, ED-TCN, TricorNet and TDRN models, we model the temporal context as a separate information stream and effectively determine the flow of information from historical embeddings through gated attention units. We believe this enables the proposed  \textit{SSA-GAN} model to oversee the evolution of sub-actions and the relationships between them more effectively. 

When comparing the results obtained from our proposed \textit{SSA-GAN} model for three datasets, the MERL dataset has the higher results as it contains only five actual action classes. Therefore, the examples for each action class during training are also higher. The Georgia Tech egocentric dataset has the lowest performance compared to the other datasets. Egocentric datasets are composed of videos that are obtained through head mounted cameras worn by the subjects on a baseball cap. Hence, these videos have a high degree of variation as the video characteristics are based on the camera wearer and their head movements. Further, the MERL Shopping and 50 Salads datasets contain similar environmental settings through out the dataset; but the Georgia Tech dataset has varied environmental settings which means the model has to learn a representation that is invariant to the environment, which is a harder task.    

\begin{table}[!ht]
\centering
\resizebox{1\linewidth}{!}{
\begin{tabular}{|p{3cm}|c|c|c|c|c|c|}
\hline
Dataset                                  & Method            & Approach                                   & F1@\{10,25,50\}                                   & edit                                  & mAP@mid                               & accuracy                              \\ \hline
                                &  \multirow{2}{*}{ Non-Deep Learning}                       & IDT+LM \cite{IDT_LM}                                    & 44.4, 38.9, 27.8                                  & 45.8                                  & NA                                    & 48.7                                  \\ \cline{3-7} 
 				&		& IDT+CRF \cite{rupprecht2016sensor}                                    & NA                                 & NA                                 & NA                                    & 54.28                                  \\ \cline{2-7} 
 
                                 & \multirow{7}{*}{ Deep-Non-GAN}                    & Spatial CNN \cite{lea2016seg}                                & 32.3, 27.1, 18.9                                  & 24.8                                  & NA                                    & 54.9                                  \\ \cline{3-7} 
                                  &                   & Dilated TCN \cite{leaCVPR}                                & 52.2, 47.6, 37.4                                  & 43.1                                  & NA                                    & 59.3                                  \\ \cline{3-7} 
                                  &                   & ST-CNN \cite{lea2016seg}                                     & 55.9, 49.6, 37.1                                  & 45.9                                  & NA                                    & 59.4                                  \\ \cline{3-7} 
                                  &                & Bi-LSTM \cite{leaCVPR}                                   & 62.6, 58.3, 47.0                                  & 55.6                                  & NA                                    & 55.7                                  \\ \cline{3-7} 
                                  &                & ED-TCN \cite{leaCVPR}                                    & 68.0, 63.9, 52.6                                  & 59.8                                  & NA                                    & 64.7                                  \\ \cline{3-7} 
                                  &                & TricorNet \cite{tricornet}                                    & 70.1, 67.2, 56.6                                  & 62.8                                  & NA                                    & 67.5                                  \\ \cline{3-7} 
                                  &                 & TDRN \cite{TDRN2018}                                    & 72.9, 68.5, 57.2                                  & 66.0                                  & NA                                    & 68.1                                  \\ \cline{2-7} 

\multirow{-7}{3cm}{50 Salads \cite{50salads}}    & GAN based    & \cellcolor[HTML]{C0C0C0}\textit{SSA-GAN} & \cellcolor[HTML]{C0C0C0}\textbf{74.9, 71.7, 67.0} & \cellcolor[HTML]{C0C0C0}\textbf{69.8} & \cellcolor[HTML]{C0C0C0}\textbf{71.4}     & \cellcolor[HTML]{C0C0C0}\textbf{73.3} \\ \hline \hline

                       &  \multirow{4}{*}{ Deep-Non-GAN}          & MSN Det \cite{merlshopping}                                   & 46.4, 42.6, 25.6                                  & NA                                    & 81.9                                  & 64.6                                  \\ \cline{3-7} 
                       &                              & MSN Seg \cite{merlshopping}                                   & 80.0, 78.3, 65.4                                  & NA                                    & 69.8                                  & 76.3                                  \\ \cline{3-7} 
                        &                             & Dilated TCN \cite{leaCVPR}                               & 79.9, 78.0, 67.5                                  & NA                                    & 75.6                                  & 76.4                                  \\ \cline{3-7} 
                      &                               & ED- TCN \cite{leaCVPR}                                   & 86.7, 85.1, 72.9                                  & NA                                    & 74.4                                  & 79.0                                  \\ \cline{2-7} 
 \multirow{-5}{3cm}{MERL Shopping \cite{merlshopping}}    & GAN based                    & \cellcolor[HTML]{C0C0C0}\textit{SSA-GAN} & \cellcolor[HTML]{C0C0C0}\textbf{92.4, 88.3, 84.3} & \cellcolor[HTML]{C0C0C0}\textbf{89.4}     & \cellcolor[HTML]{C0C0C0}\textbf{90.7} & \cellcolor[HTML]{C0C0C0}\textbf{92.1} \\ \hline \hline
 
 &  \multirow{3}{*}{ Non-Deep Learning}                       & STIP \cite{laptev2005space}  $+$ SVM                                  & NA                                  & NA                                  & NA                                    & 14.4                                  \\ \cline{3-7} 
 				&		& SIFT \cite{lowe2004distinctive}  $+$ SVM                                  & NA                                 & NA                                 & NA                                    & 29.1                                  \\ \cline{3-7} 
				&		& HO \cite{gtea_evalpaper} $+$ SVM                                     & NA                                 & NA                                 & NA                                    & 47.7                                  \\ \cline{2-7} 
				
                      &  \multirow{8}{*}{ Deep-Non-GAN}                               & EgoNet+TDD \cite{singh2016Egocentric}                                & NA                                                & NA                                    & NA                                    & 64.4                                  \\ \cline{3-7} 
                                  &                   & Spatial CNN \cite{lea2016seg}                                & 41.8, 36.0, 25.1                                  & NA                                    & NA                                    & 54.1                                  \\ \cline{3-7} 
                               &                      & ST-CNN \cite{lea2016seg}                                      & 58.7, 54.4, 41.9                                  & NA                                    & NA                                    & 60.6                                  \\ \cline{3-7} 
                                 &                    & Dilated TCN \cite{leaCVPR}                                & 58.8, 52.2, 42.2                                  & NA                                    & NA                                    & 58.3                                  \\ \cline{3-7} 
                                  &                   & Bi-LSTM \cite{leaCVPR}                                     & 66.5, 59.0, 43.6                                  & NA                                    & NA                                    & 58.3                                  \\ \cline{3-7} 
                                  &                   & ED- TCN \cite{leaCVPR}                                    & 72.2, 69.3, 56.0                                  & NA                                    & NA                                    & 64.0                                  \\ \cline{3-7} 
                                   &                   & TricorNet \cite{tricornet}                                  & 76.0, 71.1, 59.2                                  & NA                                    & NA                                    & 64.8                                  \\ \cline{3-7} 
                                   &                    & TDRN \cite{TDRN2018}                                   & 79.2, 74.4, 62.7                                  & 74.1                                   & NA                                    & 70.1                                  \\ \cline{2-7} 
\multirow{-7}{3cm}{Georgia Tech Egocentric Activities \cite{GTEA_1}} & GAN based  & \cellcolor[HTML]{C0C0C0}\textit{SSA-GAN} & \cellcolor[HTML]{C0C0C0}\textbf{80.6, 79.1, 74.2} & \cellcolor[HTML]{C0C0C0}\textbf{76.0}     & \cellcolor[HTML]{C0C0C0}\textbf{73.9}     & \cellcolor[HTML]{C0C0C0}\textbf{74.4} \\ \hline
\end{tabular}}
\caption{ Action segmentation results for 50 Salads, MERL Shopping and Georgia Tech Egocentric Activities datasets : F1@k is the segmental F1 score, edit is the segmental edit score metric (see \cite{lea2016}), mAP@mid is the mean average precision with mid point hit criterion and accuracy denotes the frame wise accuracy. NA indicates that the metric is unavailable in the respective baseline method.}
\label{tab:tab_new}
\end{table}

\begin{figure*}[!ht]
 \centering
 \includegraphics[width=.85\textwidth]{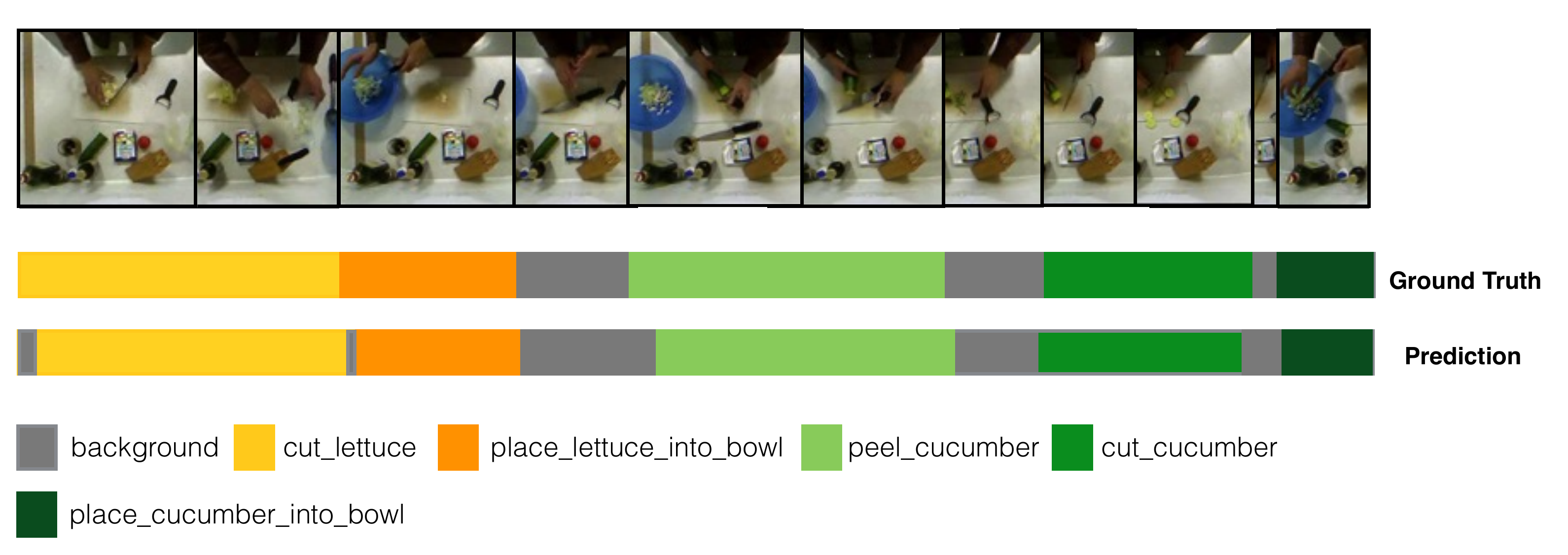} 
 \includegraphics[width=.85\textwidth]{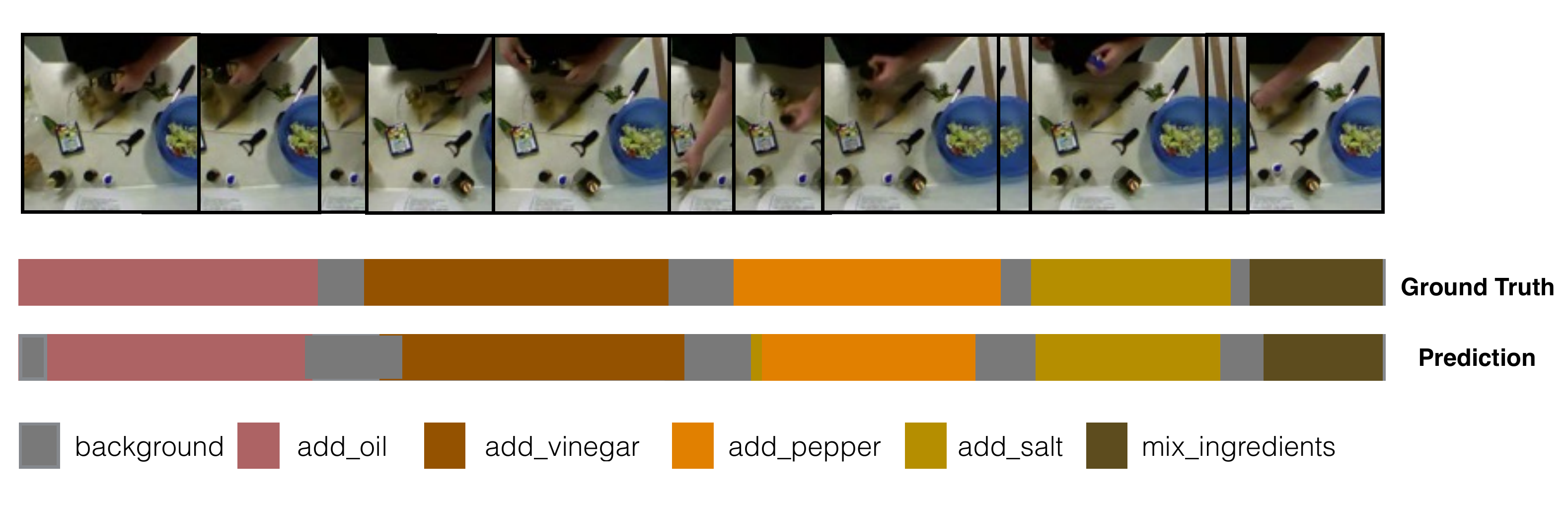} 
  \vspace{-5mm}
 \caption{Prediction results of the proposed \textit{SSA-GAN} for the 50 Salads dataset \cite{50salads}.}
\label{fig:result_3}
\end{figure*}

\begin{figure*}[!ht]
 \centering
 \includegraphics[width=.85\textwidth]{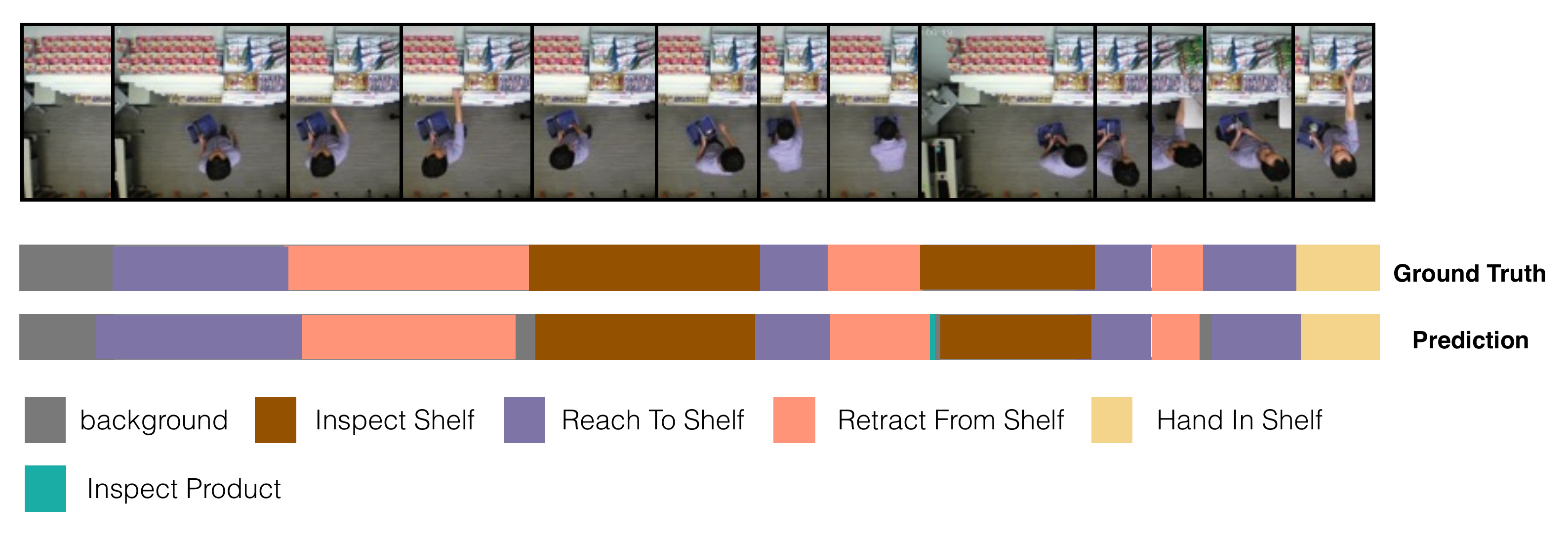} 
 \includegraphics[width=.85\textwidth]{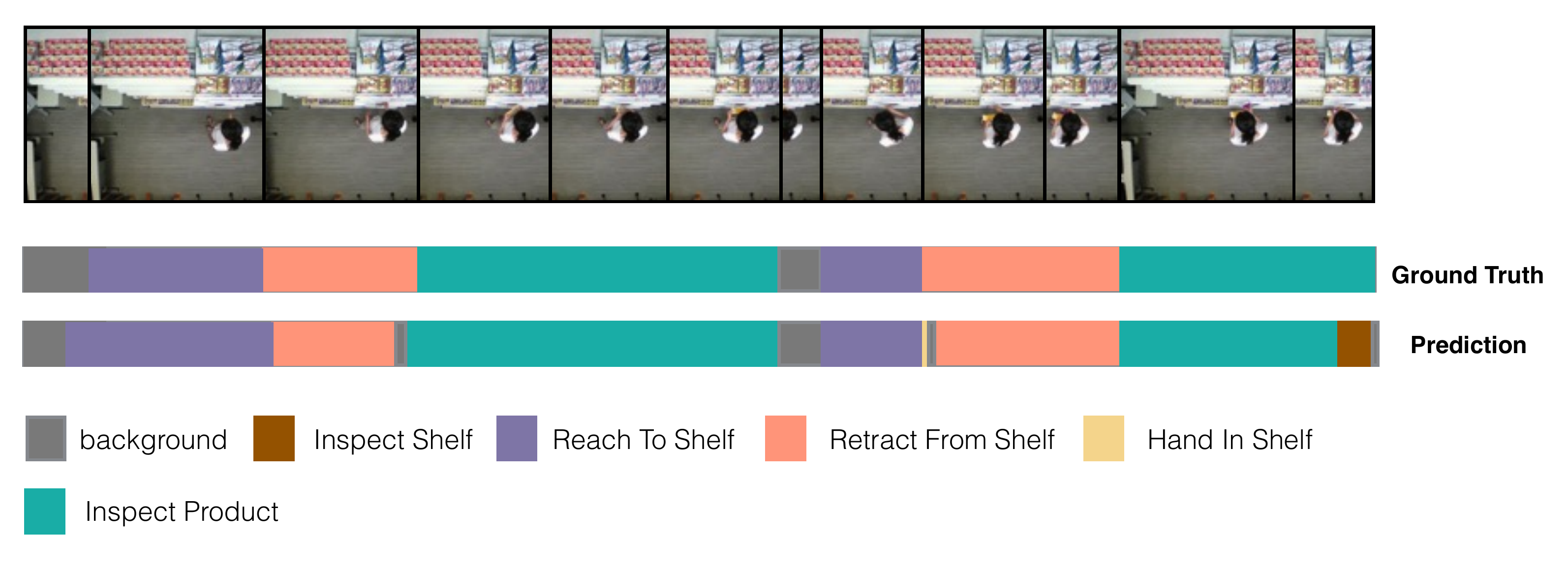} 
  \vspace{-5mm}
 \caption{Prediction results of the proposed \textit{SSA-GAN} for the MERL Shopping dataset \cite{merlshopping}.}
\label{fig:result_4}
\end{figure*}

\begin{figure*}[!ht]
 \centering
 \includegraphics[width=.85\textwidth]{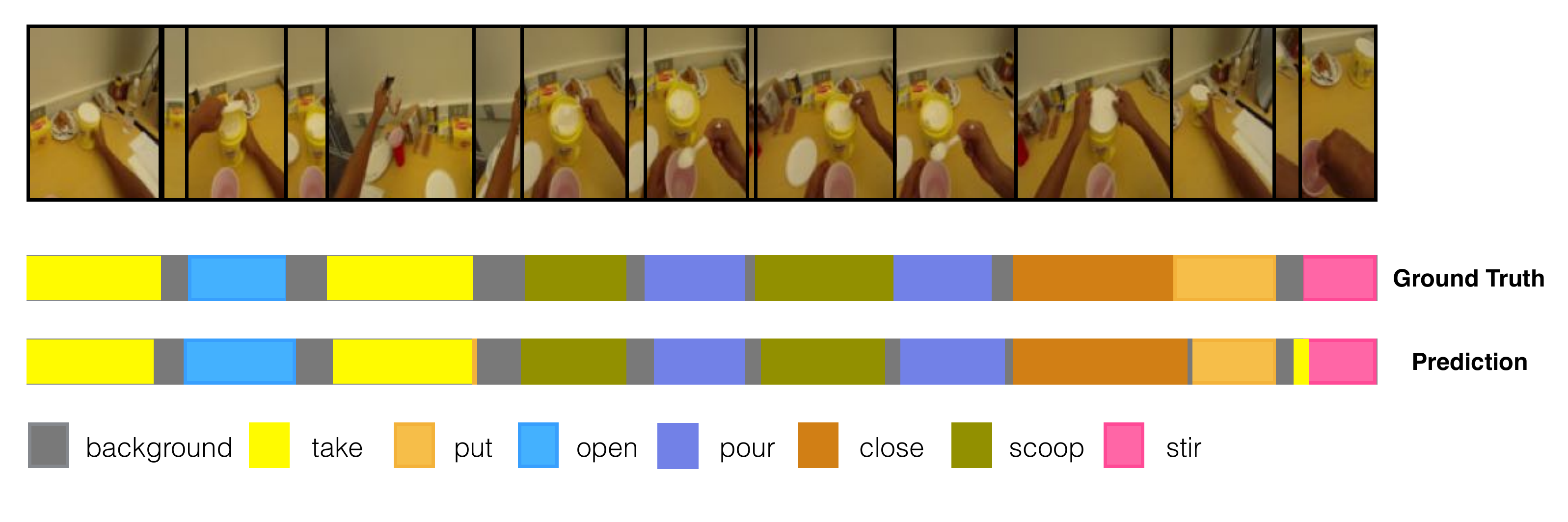} 
 \includegraphics[width=.85\textwidth]{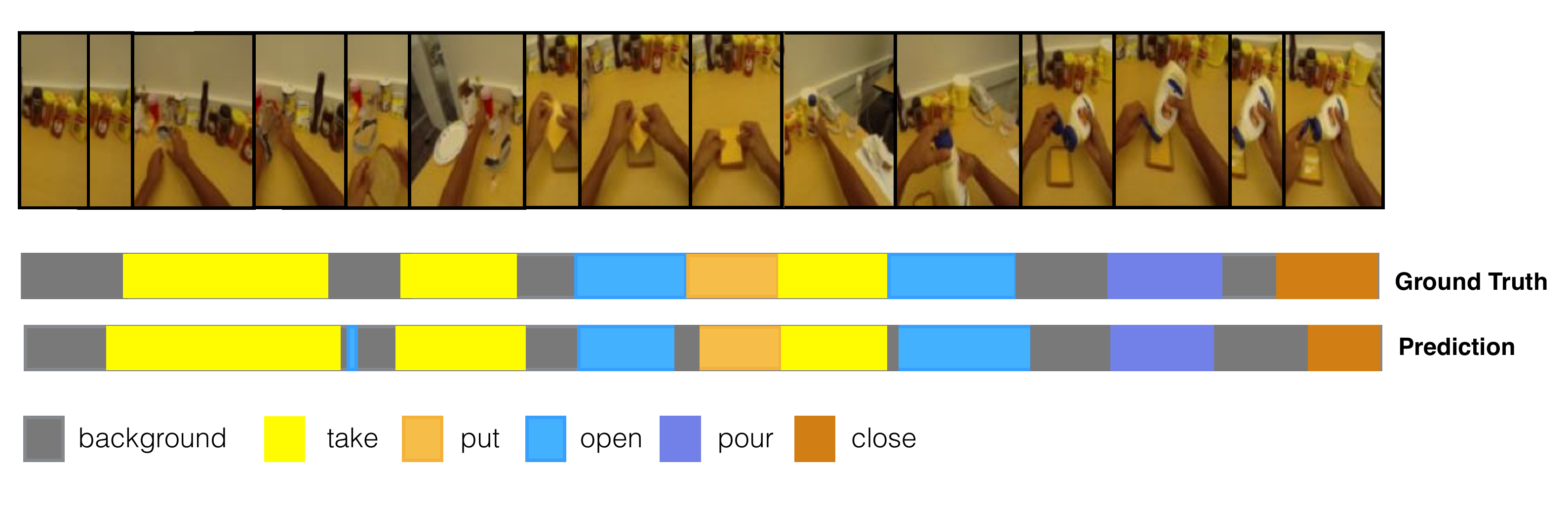} 
  \vspace{-5mm}
 \caption{Prediction results of the proposed \textit{SSA-GAN} for the Georgia Tech Egocentric Activities dataset \cite{GTEA_1}.}
\label{fig:result_5}
\end{figure*}

Figures \ref{fig:result_3} , \ref{fig:result_4} and \ref{fig:result_5} further demonstrate the performance of the proposed approach by showing the predictions against the ground truth labels in different video streams for the 50 Salads \cite{50salads}, MERL Shopping \cite{merlshopping} and Georgia Tech Egocentric Activities datasets \cite{GTEA_1} respectively.

\subsection{Ablation Experiment}

We perform ablative experiments on the MERL Shopping dataset (selected due to the moderate dataset size) to justify the importance of each component of the proposed  \textit{SSA-GAN} architecture. We utilise five simplified models, obtained by removing components from the proposed \textit{SSA-GAN} model.

\begin{enumerate}[label=\arabic*)]
\item \textbf{\textit{G-GCE}} : generator architecture from \textit{SSA-GAN} and trained to predict action classes for each input frame by adding a final softmax layer. This model does not use the information from the GCE stream and is trained in a fully supervised manner with categorical cross-entropy loss. 
\vspace{-2mm}
\item \textbf{\textit{G}} : \textit{SSA-GAN} Generator with GCE, trained in a supervised manner.
\vspace{-2mm}
\item \textbf{\textit{cGAN-GCE}} : conditional GAN, optimising the objective defined in Equation \ref{eq:3}, and not using the GCE. To generate the respective classifications, similar to \textit{G-GCE} we add a softmax layer to the end of the generator.
\vspace{-2mm} 
\item \textbf{\textit{cGAN}} : \textit{SSA-GAN} model without the semi-supervised objective, coupled with the GCE. 
\vspace{-2mm}

\item $\eta=$ \textbf{\textit{(SSA-GAN)-GCE}} : semi supervised \textit{SSA-GAN} architecture optimising Equation \ref{eq:4}, although without the GCE features.

\item $\eta+$ \textbf{\textit{LSTM}} :  semi supervised \textit{SSA-GAN} architecture with an LSTM network replacing the GCE module.
\vspace{-2mm}
\end{enumerate}

Analysing the results in Table \ref{tab:ablation}, we observe a significantly lower accuracy for model \textit{G}. It is a non-GAN method inheriting the deficiencies of supervised training of CNNs, where performance is directly coupled with the design of the loss function. The introduction of temporal features boosts performance (\textit{G} compared to \textit{G-GCE}), however, it fails to achieve performance comparable to GAN based methods. Comparing \textit{cGAN-GCE} and \textit{cGAN}, we observe that context information is vital for fine-grained action segmentation. We see that encoded temporal information is capable of improving the learning process and the inclusion or exclusion of this results in a significant performance gap.

The difference between the \textit{(SSA-GAN)-GCE} and the \textit{cGAN} models is mainly due to the end-to-end action learning ability of \textit{(SSA-GAN)-GCE}. As the discriminator network of the \textit{(SSA-GAN)-GCE} model learns the classification together with the real/fake verification, it does not require any additional classification approach. The \textit{cGAN} model is trained only for action code generation and an additional softmax layer with the generator is required for action classification after the GAN based training. Therefore, it learns to generate action codes without directly learning to determine the action classes.

The SSA-GAN - GCE ablation model operates at the frame level, and due to the GAN based learning framework it has been able to optimally utilise the available spatial information and map that to an action code, which is subsequently used to classify the action class of the input frame. We believe the GAN based learning paradigm allowed us to obtain commendable accuracy for the SSA-GAN - GCE ablation model in Table \ref{tab:ablation}, while only operating on single images as single frames would carry some, but not all, information related to what action class the frame belongs to. However, as we as not using temporal information from successive frames, the introduction of the GCE allows us to obtain a substantial (approximately 5\%) accuracy increase compared to the SSA-GAN - GCE ablation model.

When comparing the results for $\eta+$\textit{LSTM} with the \textit{SSA-GAN} model, \textit{SSA-GAN} is able to outperform $\eta+$\textit{LSTM}. This is due to the fact that the LSTM network output depends only on the immediately preceding cell state. Therefore, when handling a long sequence of data with LSTMs there is higher chance of losing the long-term relationships among the samples. In contrast, the GCE module is capable of considering the entire stored sequence when determining the output at each gate. Hence, it is able to outperform the LSTM based model.     

Overall, we observe a considerable gap between models with the GCE component and those without. This is due to the attention weights of the GCE being jointly learnt with the overall model, enabling the model to understand important areas of the feature queue. As such, the \textit{SSA-GAN} model is capable of outperforming all other models.   

\begin{table}[!ht]
\begin{center}
\resizebox{0.75\linewidth}{!}{
\begin{tabular}{|c|c|c|c|}
 \hline

        Approach  &F1@\{10,25,50\} & mAP@mid & accuracy \\	
  \hline
   	 \textit{$G-GCE$}& 24.7, 23.9, 23.1  & 24.2 & 28.6  \\ 
  	 \textit{$G$}& 29.9, 27.1, 24.6 & 27.3 & 32.1 \\ 
 \hline
	 \textit{cGAN-GCE}& 77.0, 75.6, 72.7  & 74.6 & 78.1  \\ 
	 
	 \textit{cGAN}& 83.7, 83.3, 81.4 & 82.7 & 86.2 \\ 
	$\eta=$\textit{(SSA-GAN)-GCE}& 87.9, 85.8, 83.4 & 84.1 & 87.3 \\ 
	$\eta+$\textit{LSTM}& 89.8, 89.0, 83.6  & 88.9 & 90.7 \\ 
	 \textit{SSA-GAN}& \textbf{92.4}, \textbf{88.3}, \textbf{84.3} & \textbf{90.7}  & \textbf{92.1}\\ 
	
\hline
			
\end{tabular} }
\end{center}
 \vspace{-4mm}
\caption{Ablation experiment results for MERL Shopping dataset. The evaluation metrics are as defined in Section \ref{met}. Poor performance was observed in non-GAN based methods. Proposed gated context extractor and GAN based loss function learning have significantly contributed for the performance of the \textit{SSA-GAN} model.}\label{tab:ablation}
\end{table}

\section{Discussion}
\label{sec:discussion}
\subsection{Action codes}
As mentioned in Section \ref{ac_code} the action codes utilised here are just an example. They can be substituted with any representation (i.e any vectors or matrices such as images) that represents each class uniquely. The contribution between adversarial loss and classification loss is balanced using $\lambda_c$. As it is set to be 100 it provides more attention to the classification process, hence allowing the generator model to alter the action code representation to make them easily classifiable by the discriminator.  The ground truth action codes are provided as a guideline for the discriminator to guide the generator. However, by giving more weight to the classification error we place emphasis on the action codes being distinctive for recognition rather than being close to the ground truth. 

\subsection{Importance of task specific loss learning}
\begin{figure}[!ht]

\begin{center}
    \subfigure[Generator of the proposed Semi-supervised GAN model]{\includegraphics[width=.7\textwidth]{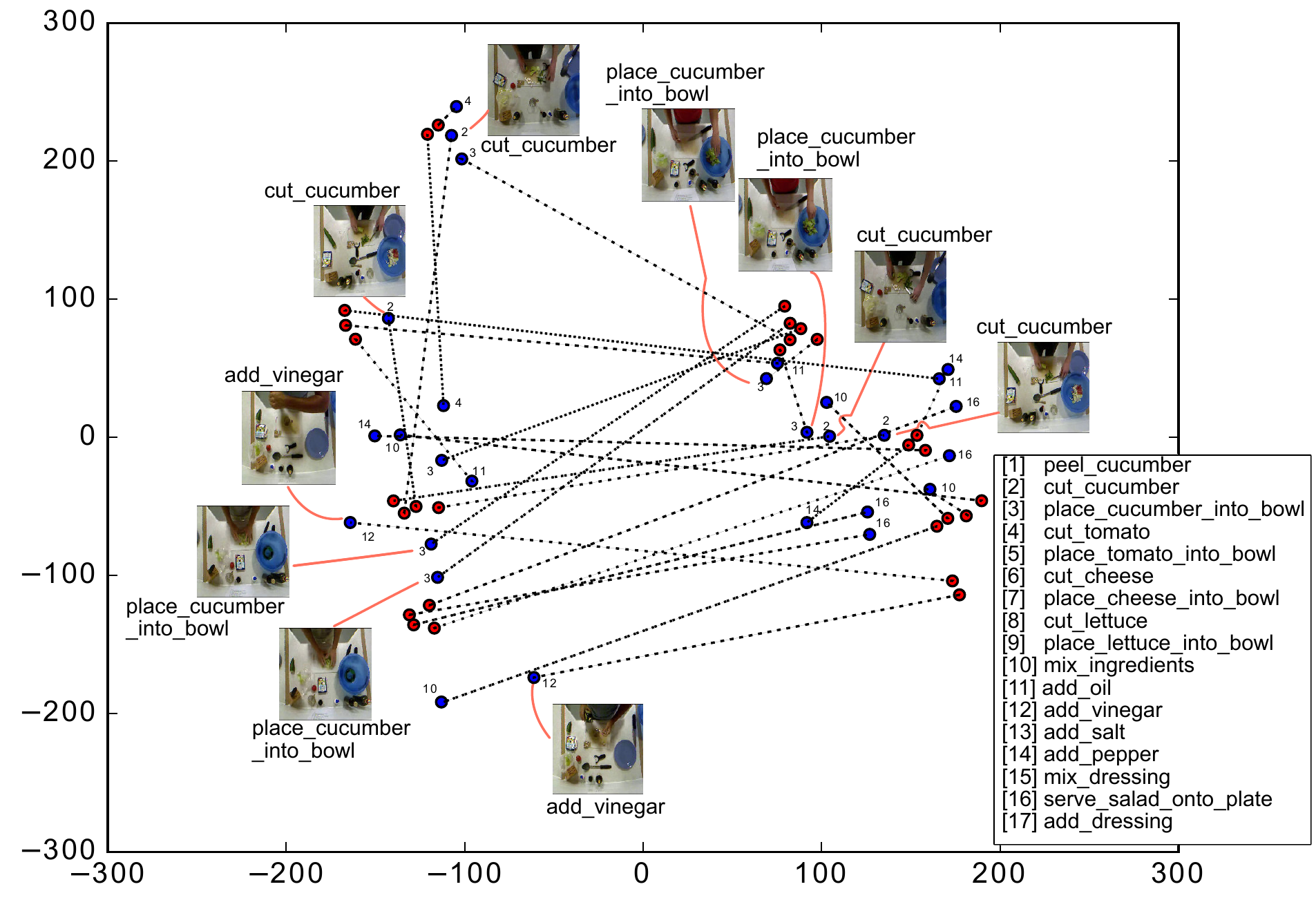}} %
     \subfigure[Supervised model (G)]{\includegraphics[width=.7\textwidth]{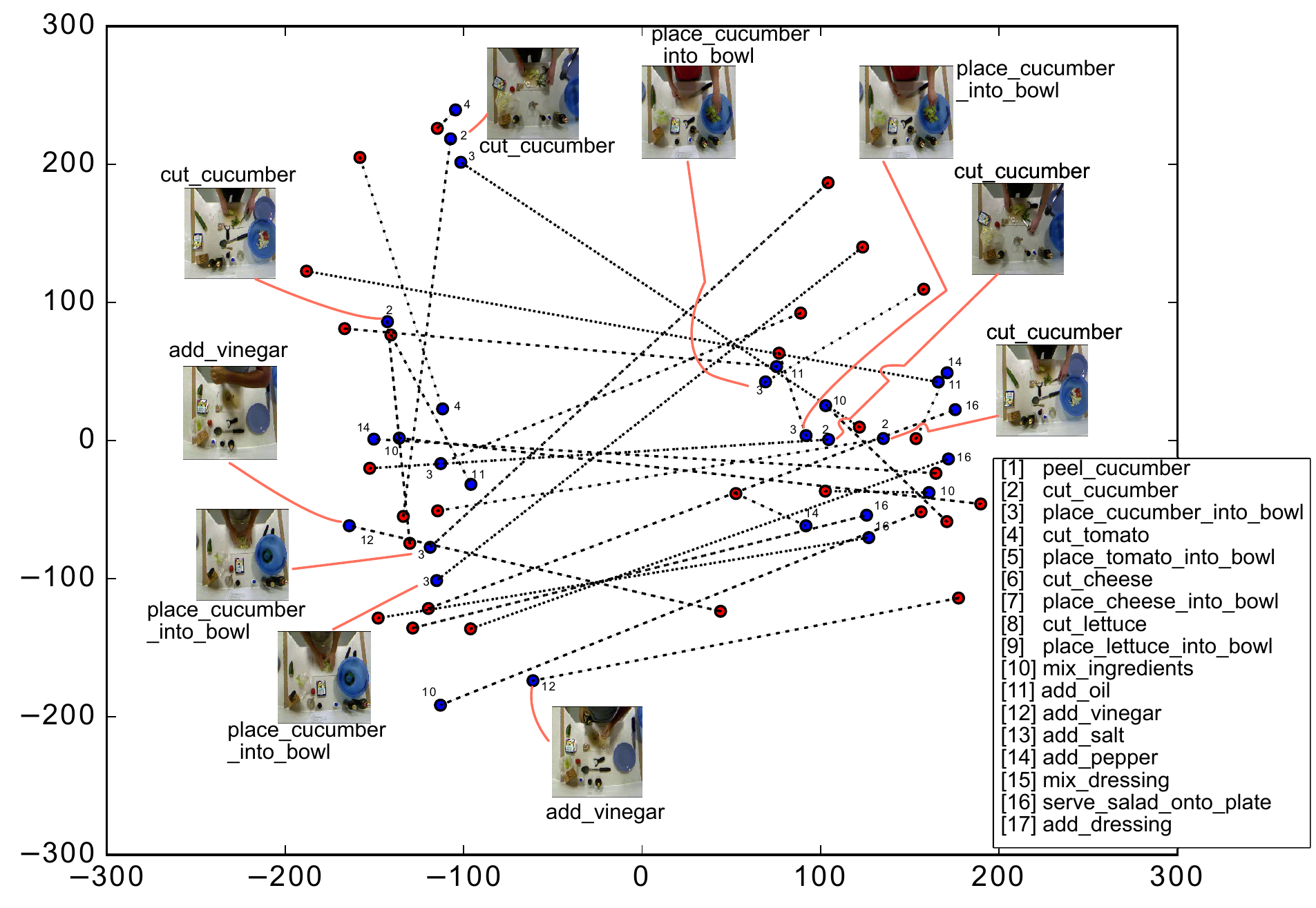}} %
 \end{center}
  \vspace{-7mm}
  \caption{Visualisation of the embedding space before (in blue) and after (in red) training.}
  \label{fig:embed_shift}
\end{figure}

We select 30 examples from the validation set of the 50 Salads dataset and Figure \ref{fig:embed_shift} (a) shows the visualisations of the embedding space positions before (in blue) and after (in red) the training of the generator model of the proposed SSA-GAN network with these examples. Similar to \cite{aubakirova2016interpreting} we extracted the activations from layer 5 and applied PCA \cite{wold1987principal} to plot them in 2D. The respective ground truth class IDs are indicated in brackets.  This provides a better understanding of the encoding process that is utilised by the generator which directs the discriminator to learn the action classification. 

With these examples, we noticed that the frames from the same action class are more tightly grouped. We also compare these semi-supervised model visualisations with a similar plot obtained for the supervised model $G$ from the ablation experiments. The 30 examples chosen from the validation set are selected as they include different subjects performing different actions. However, appearance wise all these examples inherit similar characteristics with the changes mostly occurring in subject related features such as hand and object positions. 

The embedding shift provides a visual understanding of how each sample is represented by the model after the learning process has completed. Before training the model has no idea of where to place a particular example in the embedding space. However once training completes, the examples from the same action class should be grouped together, since irrespective of the differences in input frames, the model should have learned silent action specific features to classify frames which belong to the same action class. When comparing Figure \ref{fig:embed_shift} (a) and (b) we demonstrate that when using standard cross entropy loss based learning (Figure \ref{fig:embed_shift} (b)), the model fails to achieve this. For instance, after training (denoted in red) we observe that the model has shifted samples of the place\_cucumber\_into\_bowl action class (one example shown in the bottom left corner and one on the top right side in Figure \ref{fig:embed_shift} (b)) in opposite directions. This can be observed in other action classes as well. This clearly illustrates that the model is uncertain about the action class of these examples. 												
In contrast when we analyse Figure \ref{fig:embed_shift} (a) we observe tighter grouping of the examples from the same action class. When considering the same two examples from the place\_cucumber\_into\_bowl action class, after the learning process the model has successfully shifted those two examples to the same region in the embedding space. This clearly demonstrates the discriminative learning capacity of the proposed model, where examples of the same action class are grouped together after training, which is a result of the action code based learning framework which forces the generator to utilise a task specific GAN loss to discriminate between the action classes.  

\begin{figure}[!ht]
\begin{center}
    \subfigure[Input]{\includegraphics[width=.25\textwidth, height= .195\textwidth]{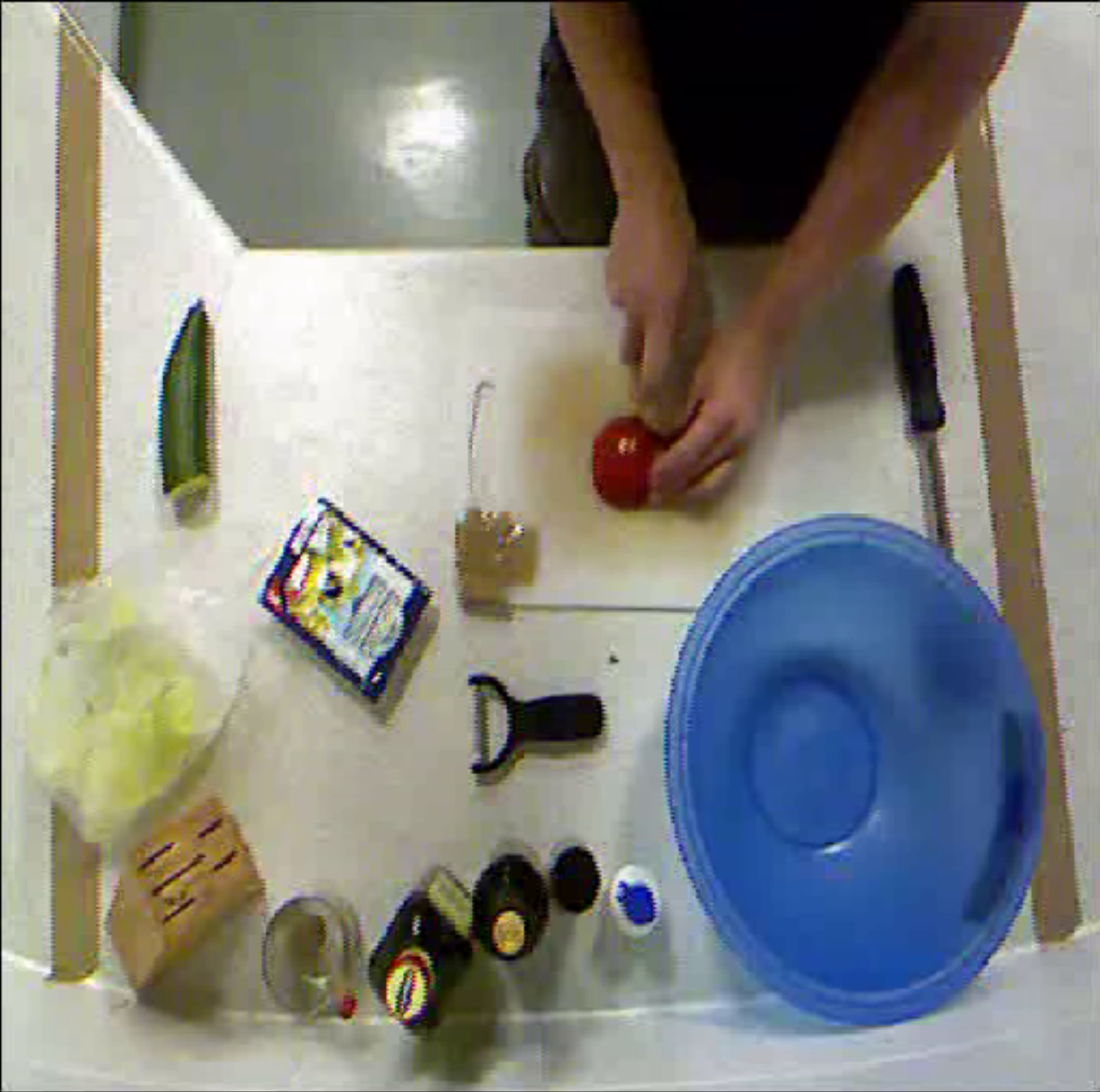}} %
     \subfigure[layer 2 activations]{\includegraphics[width=.25\textwidth]{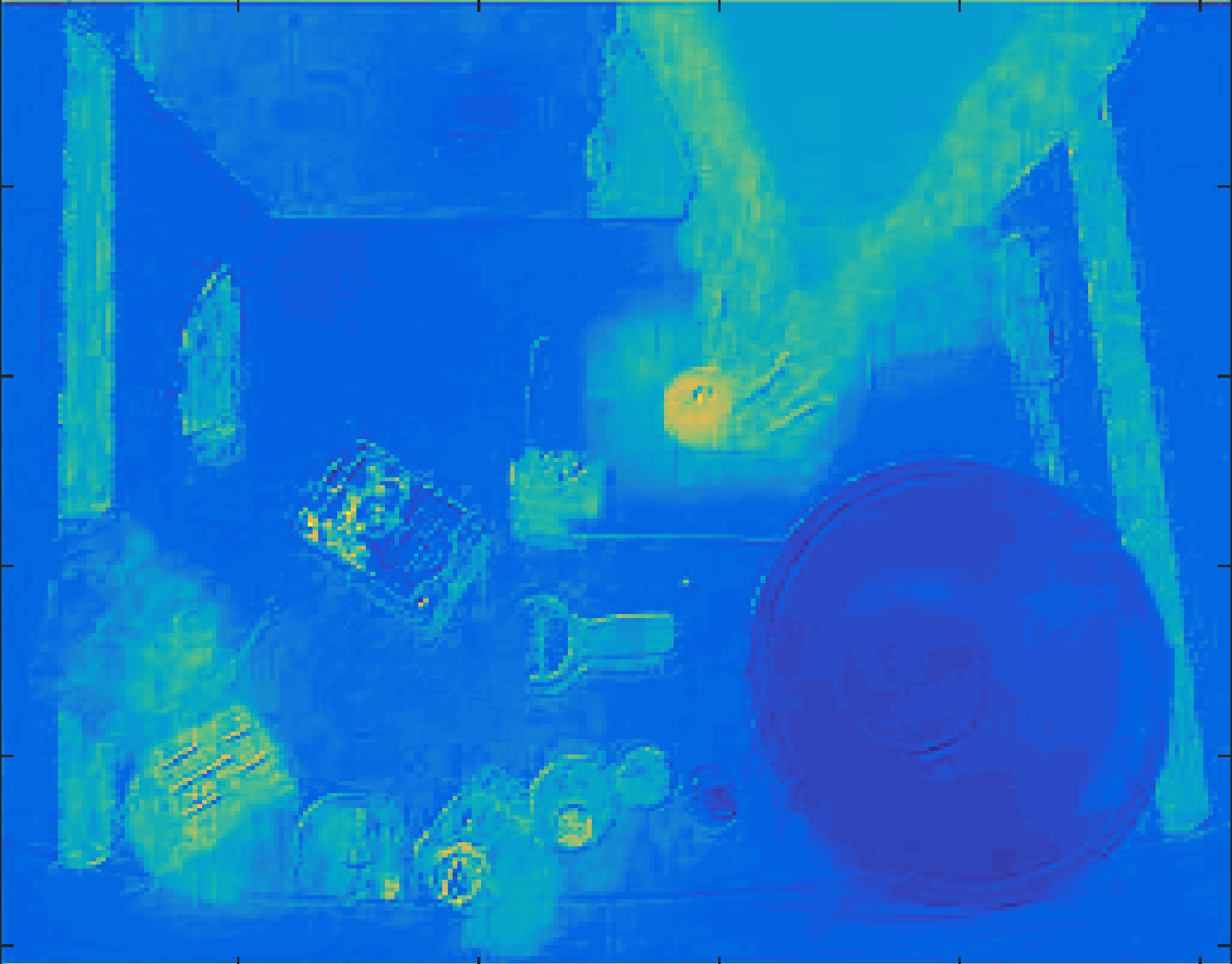}} %
       \subfigure[layer 5 activations]{\includegraphics[width=.25\textwidth]{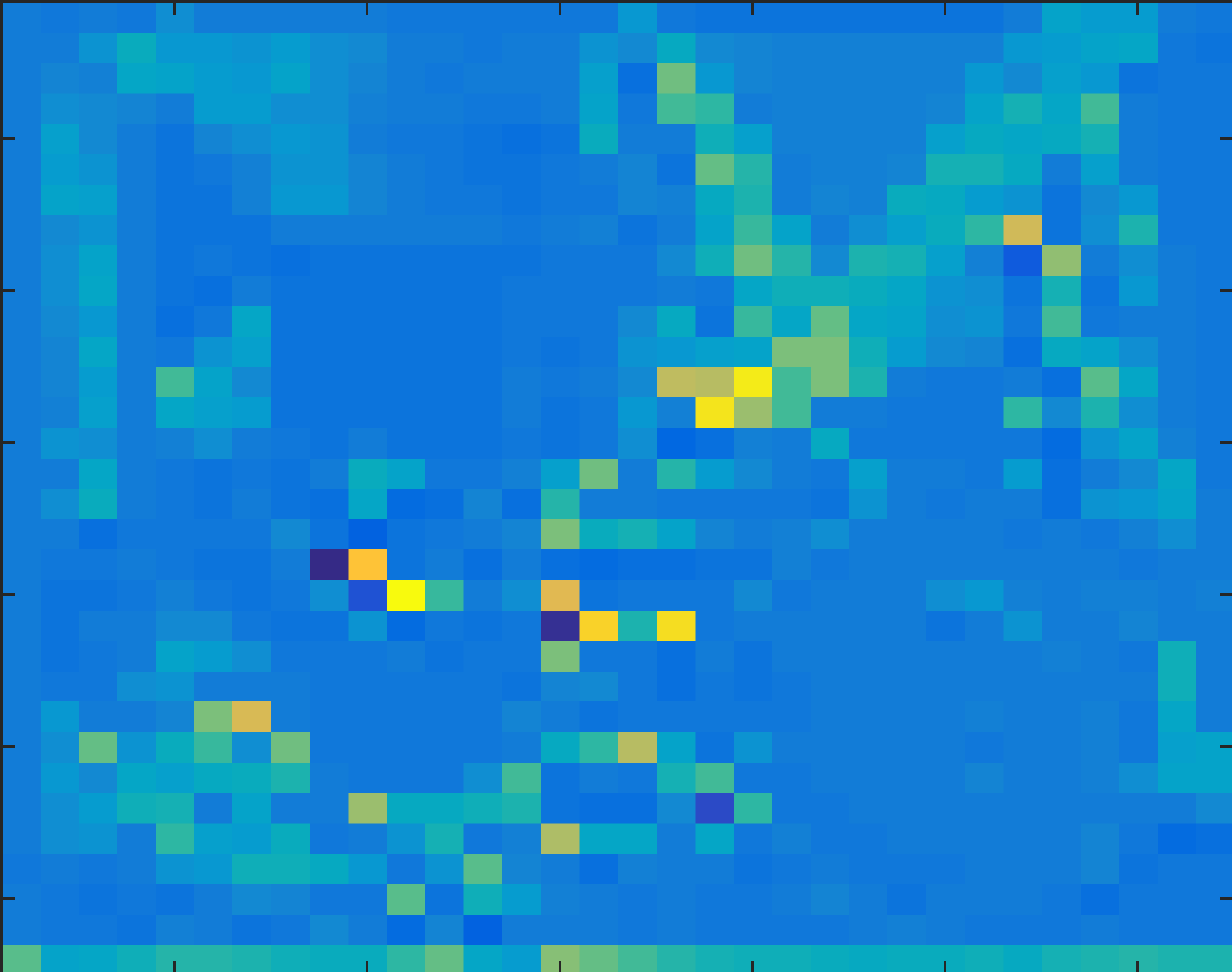}} %
 \subfigure[Input]{\includegraphics[width=.25\textwidth, height= .195\textwidth]{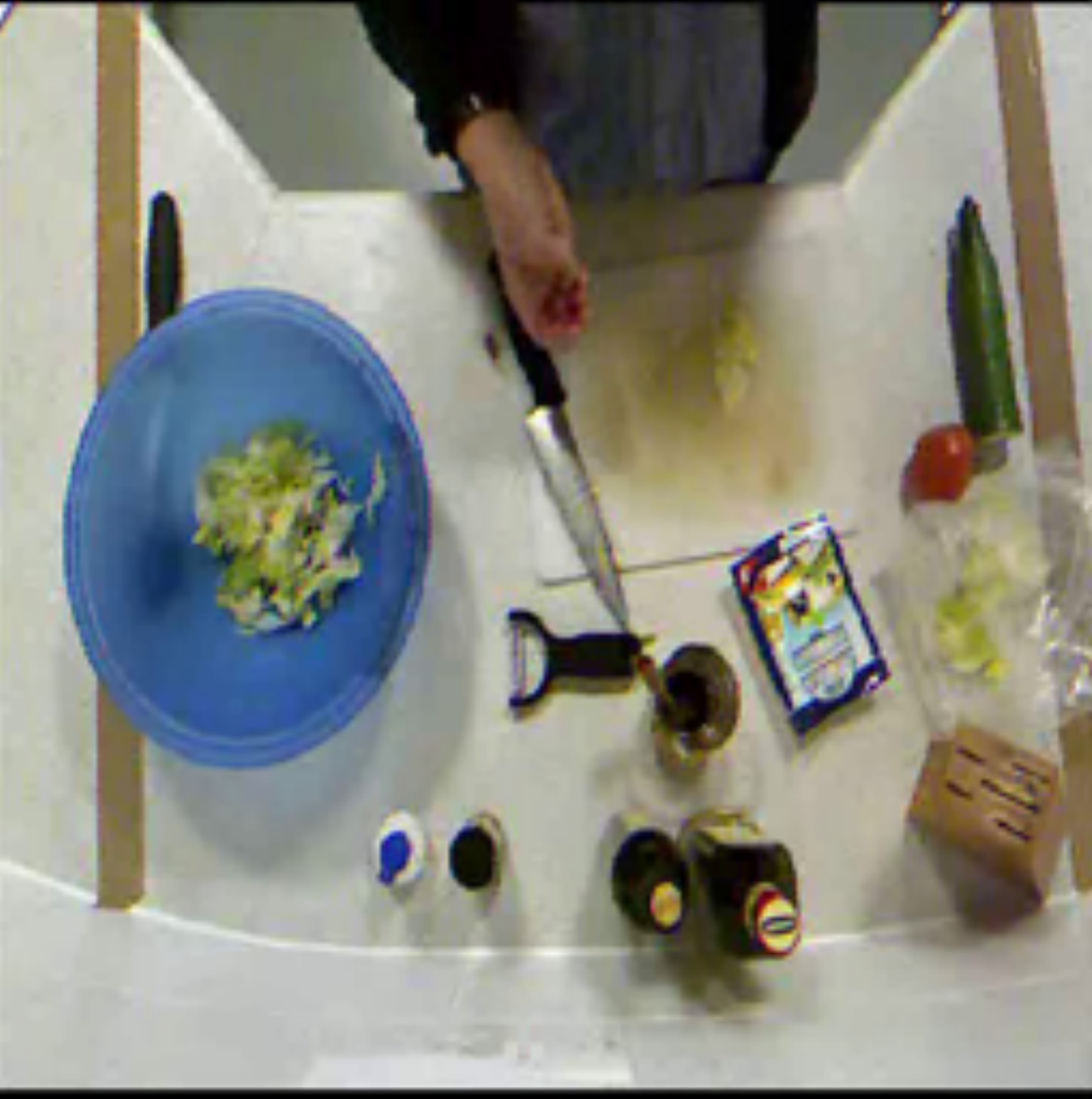}} %
     \subfigure[layer 2 activations]{\includegraphics[width=.25\textwidth]{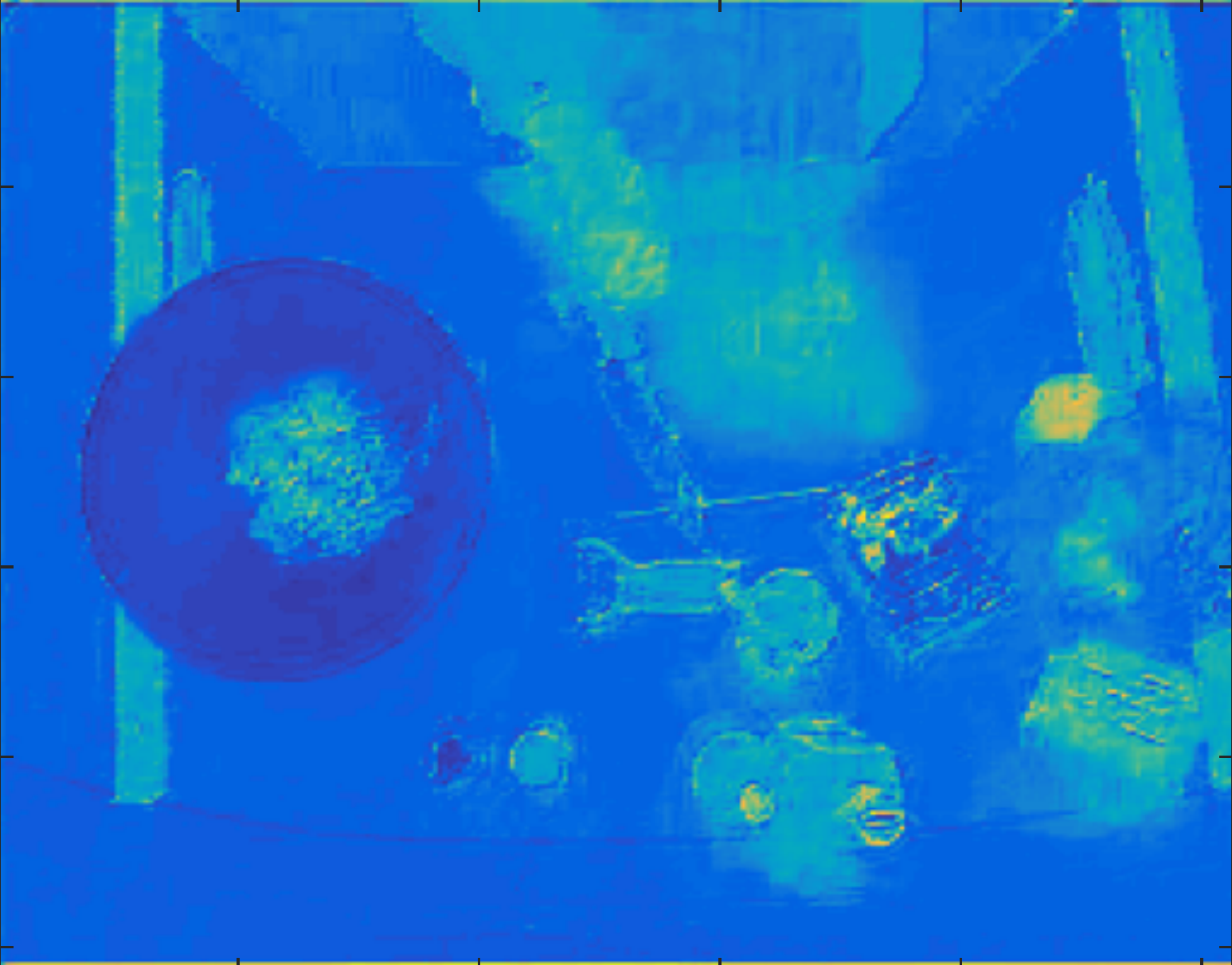}} %
       \subfigure[layer 5 activations]{\includegraphics[width=.25\textwidth]{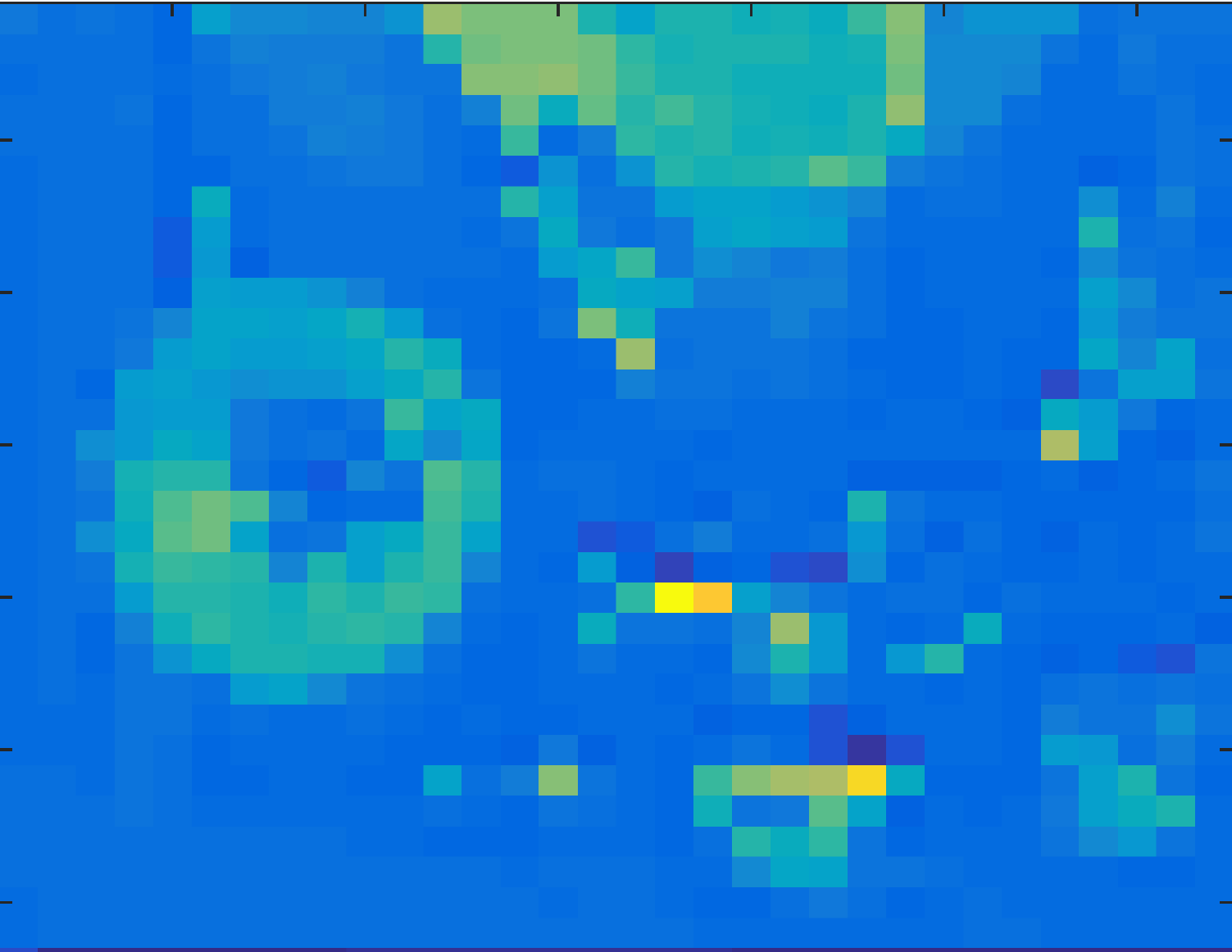}} %
 \end{center}
  \vspace{-7mm}
  \caption{Activation visualisations for different layers of the generator of the proposed SSA-GAN model. Yellow denotes more attention while blue denotes less attention.}
  \label{fig:activ_visualise}
\end{figure}

This proper grouping of related samples leads the proposed model to gain better classification results. To further examine the action recognition ability of the proposed model we visualise the trained generator model activations from the 2nd layer and the 5th layer with respect to the input image (Figure \ref{fig:activ_visualise}). In the first row, the action `cut\_tomato' mainly involves the human hand interacting with the knife and the tomato. In the early layer activations, the network gives more attention to the hand, tomato and various surrounding objects. Then in the later layer activations attention focuses more around the tomato and the hand in order to recognise the corresponding action. Similarly, in the second row of Figure \ref{fig:activ_visualise}, in order to represent the action `place\_lettuce\_into\_bowl' the model learns to give attention to the bowl, hand and other related objects. As the generator model learns an `action code' which is an intermediate representation to represent the current action, it tries to capture action specific regions and their details within the input that aid in better describing the action. 
  
\begin{figure}[!ht]
\begin{center}
    \subfigure[Input]{\includegraphics[width=.25\textwidth, height= .195\textwidth]{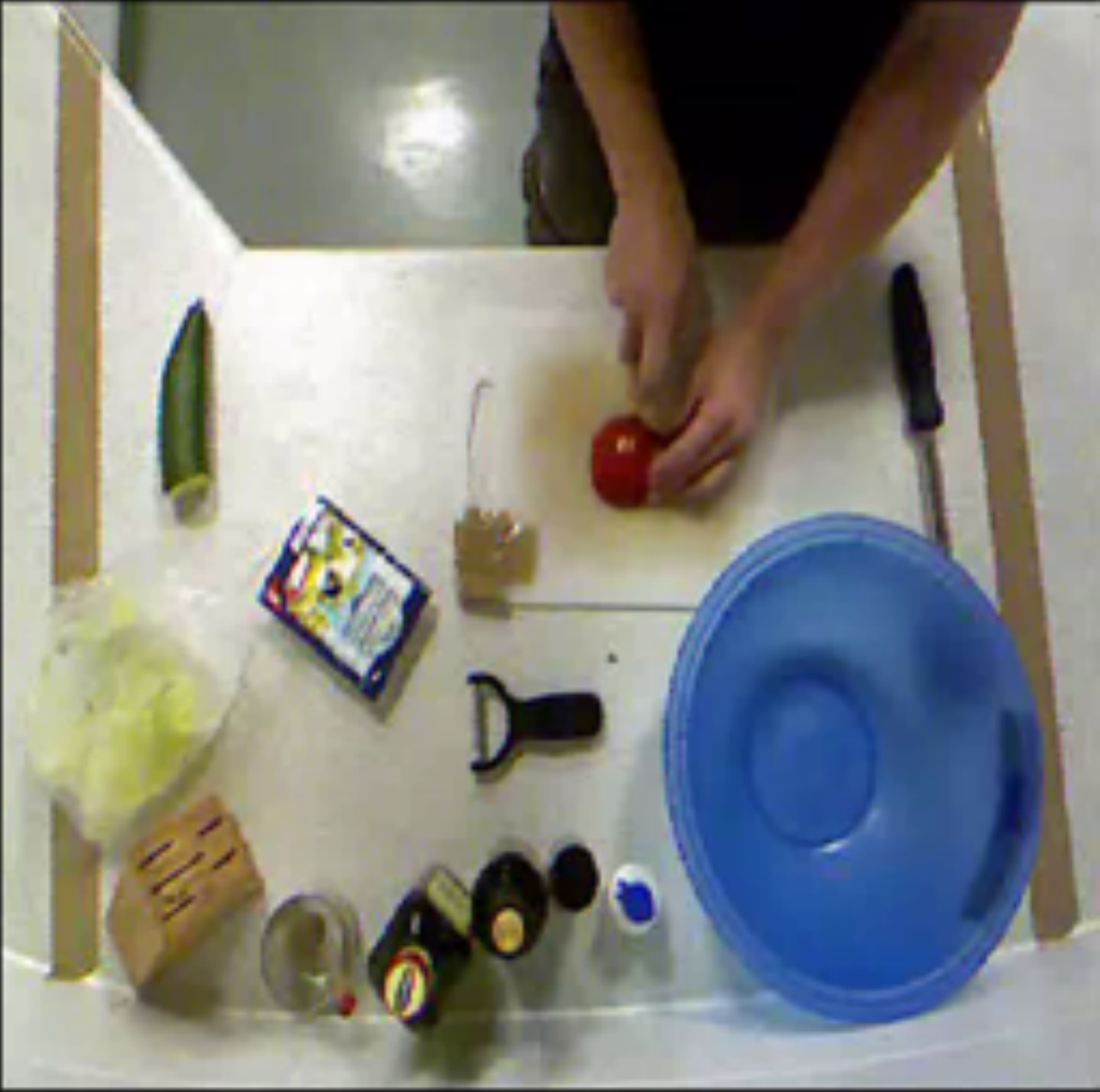}} %
     \subfigure[layer 2 activations]{\includegraphics[width=.25\textwidth]{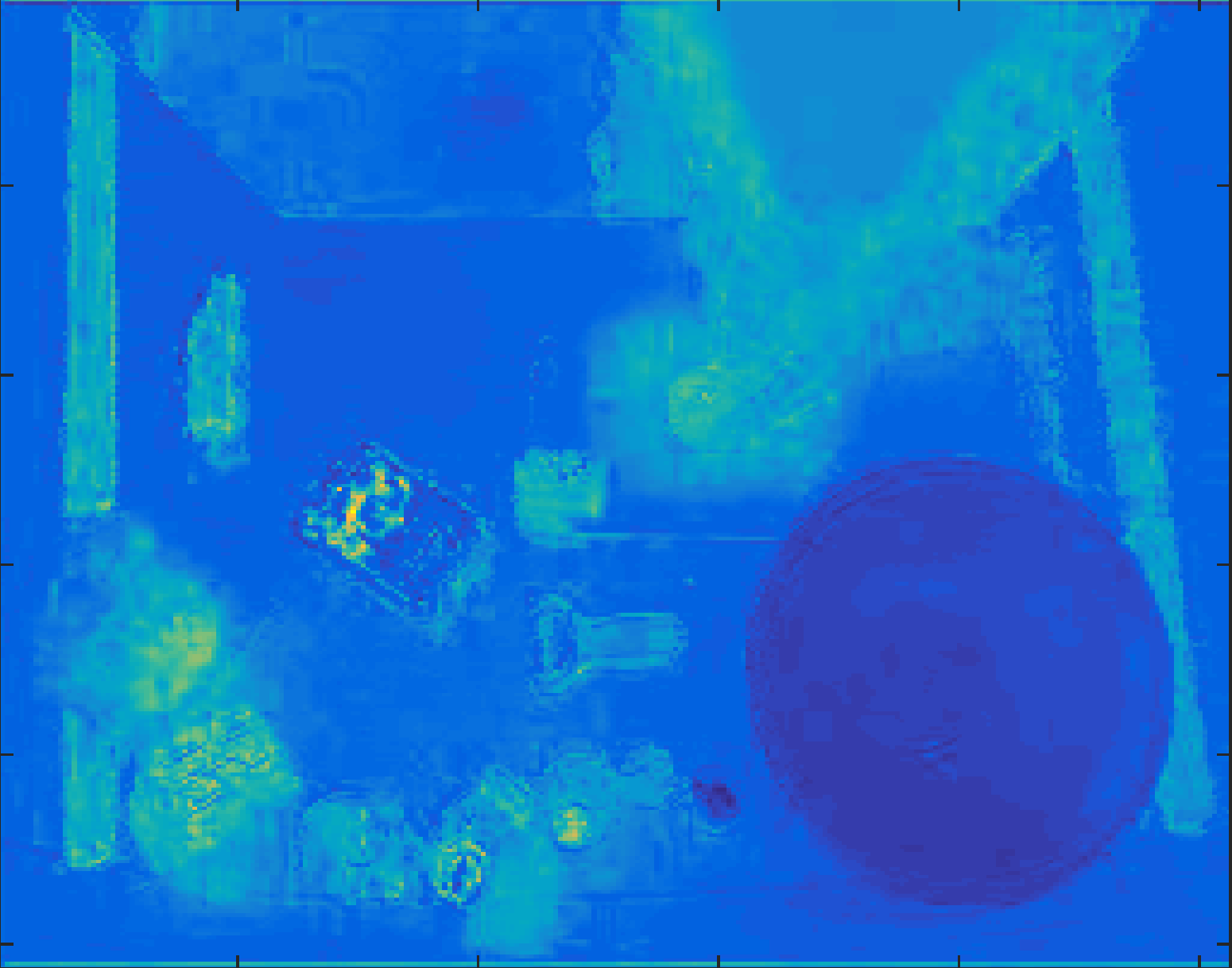}} %
       \subfigure[layer 5 activations]{\includegraphics[width=.25\textwidth]{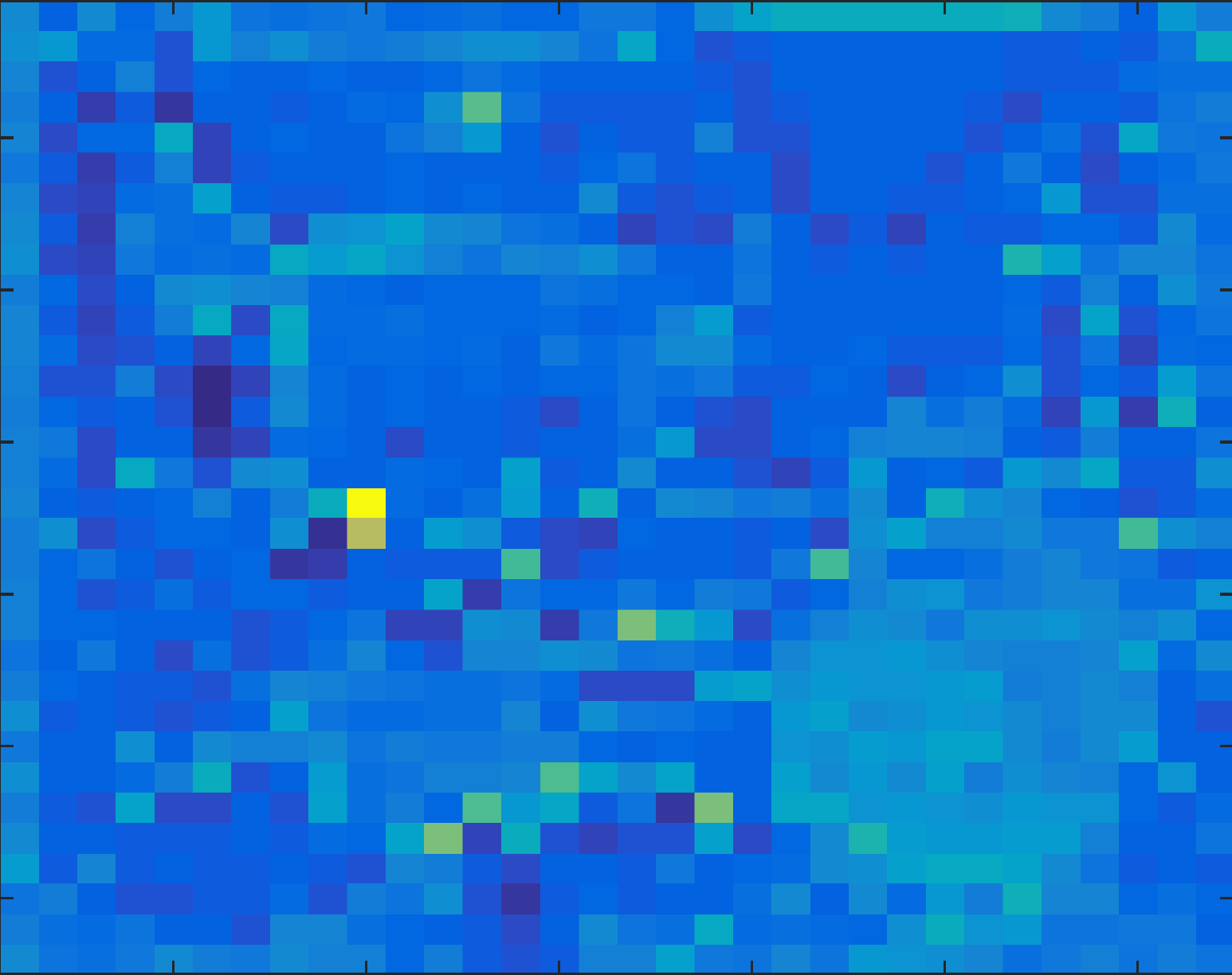}} %
 \subfigure[Input]{\includegraphics[width=.25\textwidth, height= .195\textwidth]{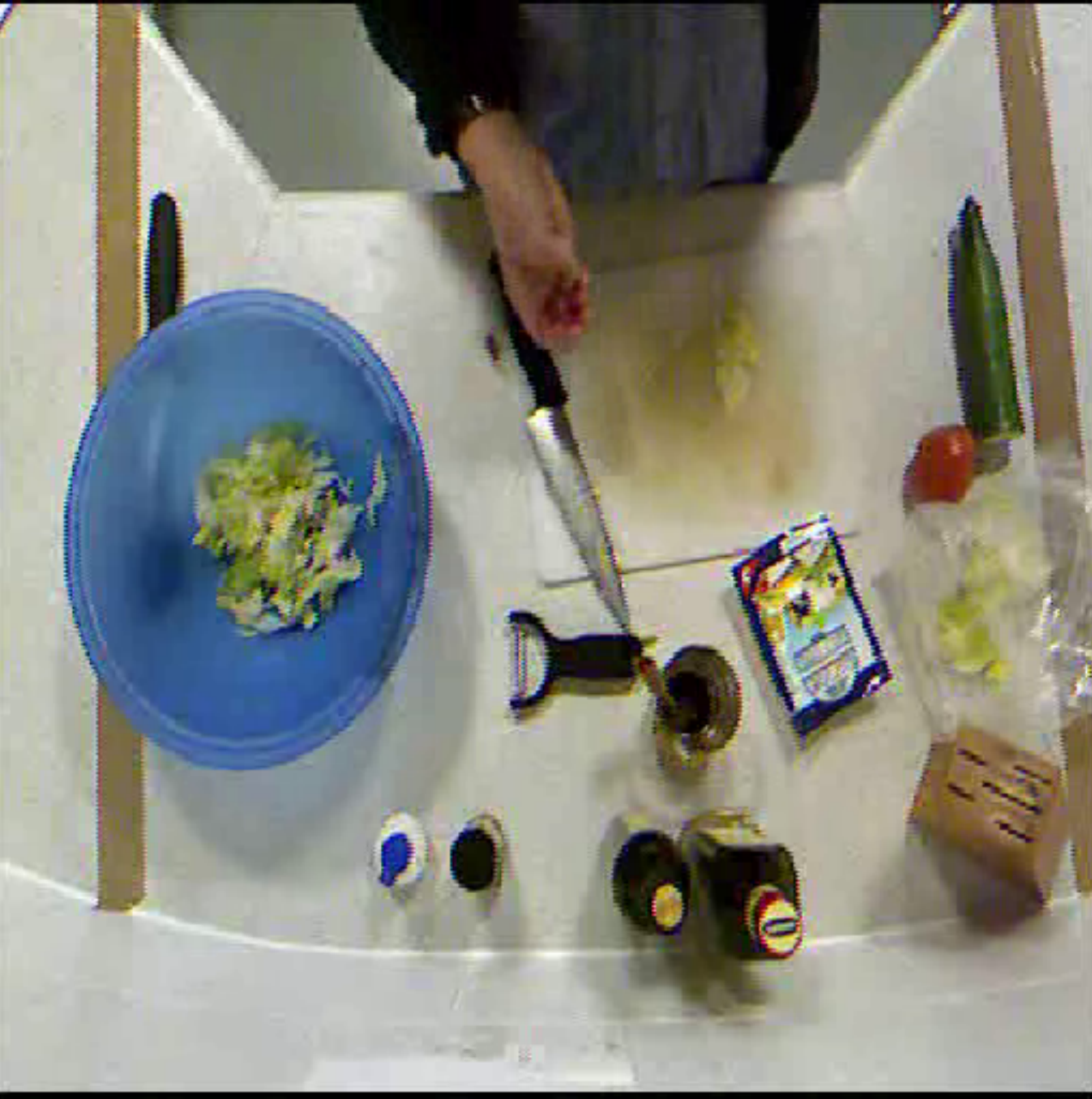}} %
     \subfigure[layer 2 activations]{\includegraphics[width=.25\textwidth]{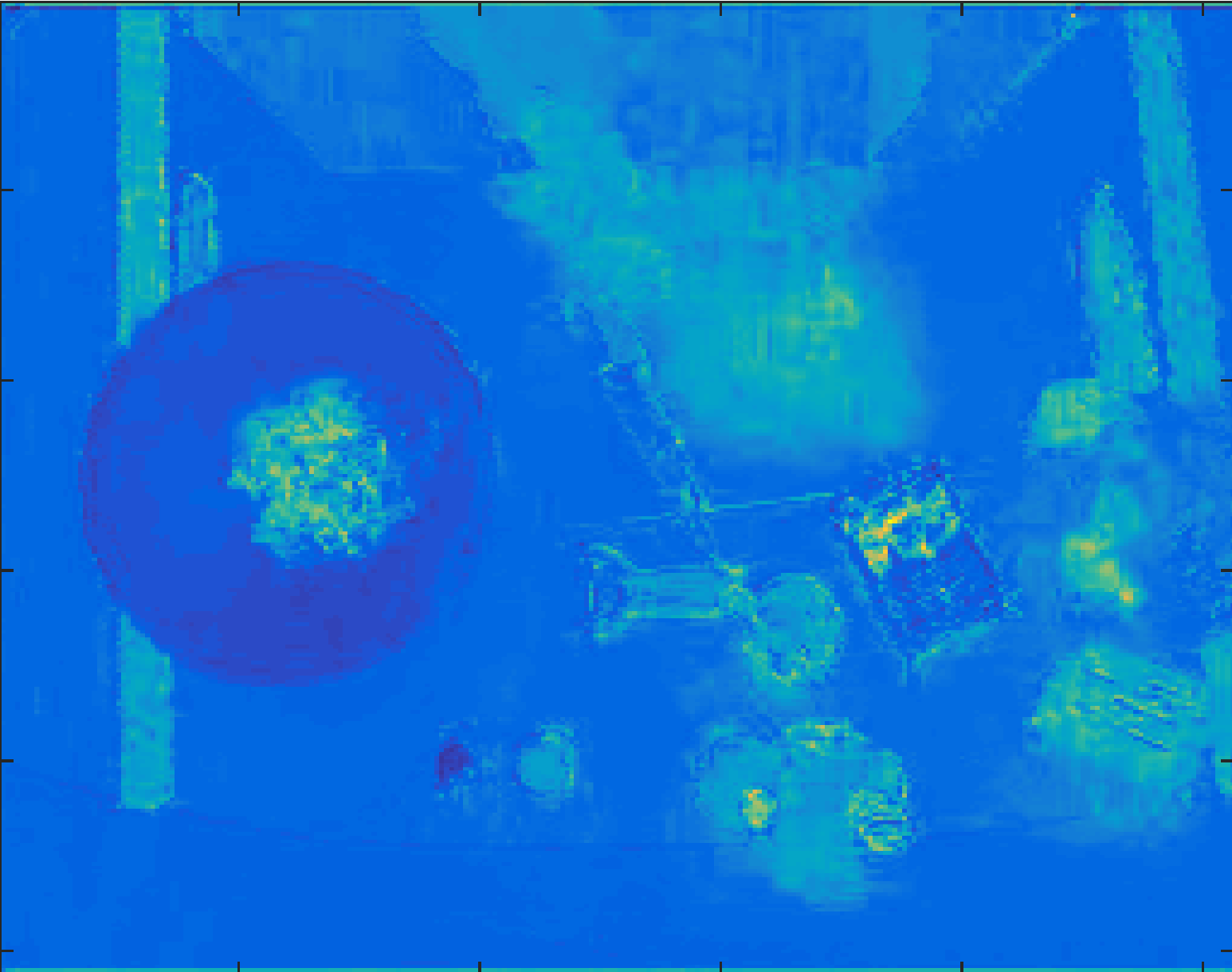}} %
       \subfigure[layer 5 activations]{\includegraphics[width=.25\textwidth]{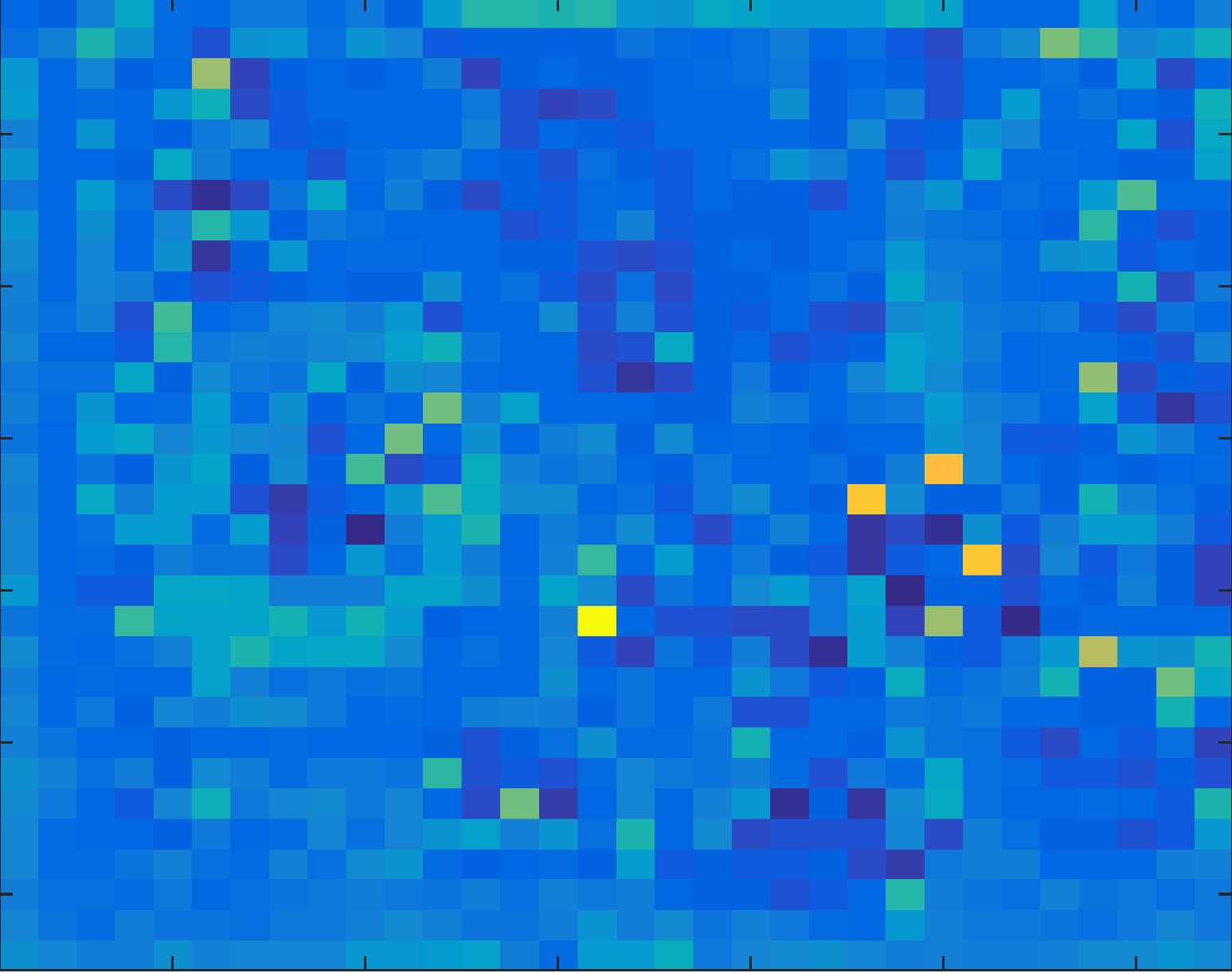}} %
 \end{center}
 \vspace{-7mm}
  \caption{Activation visualisations for different layers of the supervised model G (see ablation experiments). Yellow denotes more attention while blue denotes less attention.}
  \label{fig:activ_visualise_supervised}
\end{figure}

From the activations shown in Figure \ref{fig:activ_visualise_supervised} from the supervised model G from the ablation experiments, it is clear that the automatic feature extraction process without such guidance fails to capture salient action related information. We believe this results in the cluttered distribution of the features from the supervised model G seen in Figure \ref{fig:activ_visualise_supervised}; justifying the lack of substantial improvement among the baseline methods such as Bi-LSTM, ED-TCN and TricorNet (i.e the performance increase among any 2 models in F1 score is 5 units). This is because these models try to map these spatially similar incoming pixels directly to a classification label, using a standard supervised learning objective. 

\begin{figure}[!ht]
\begin{center}
    \subfigure[Input]{\includegraphics[height=.25\textwidth]{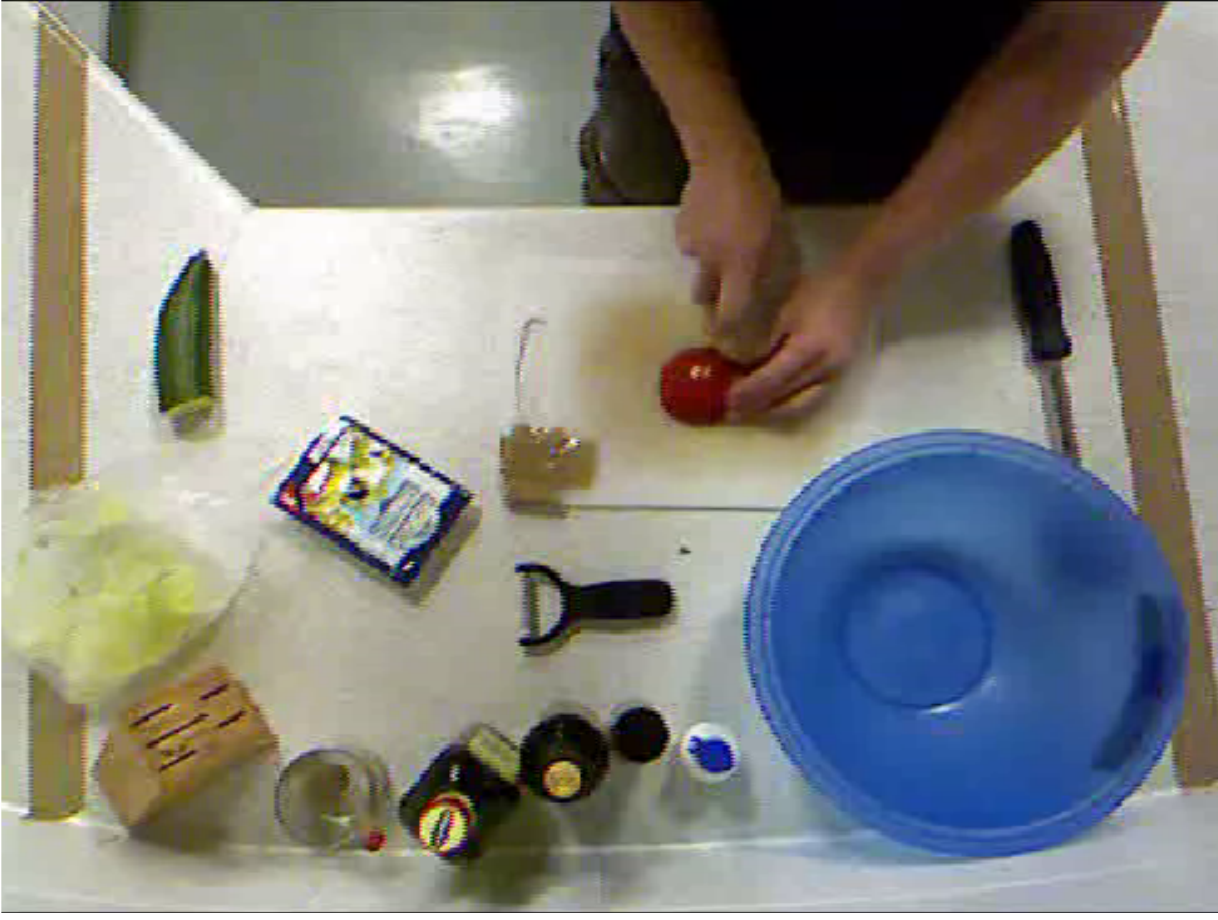}} %
     \subfigure[layer 2 activations]{\includegraphics[height=.25\textwidth]{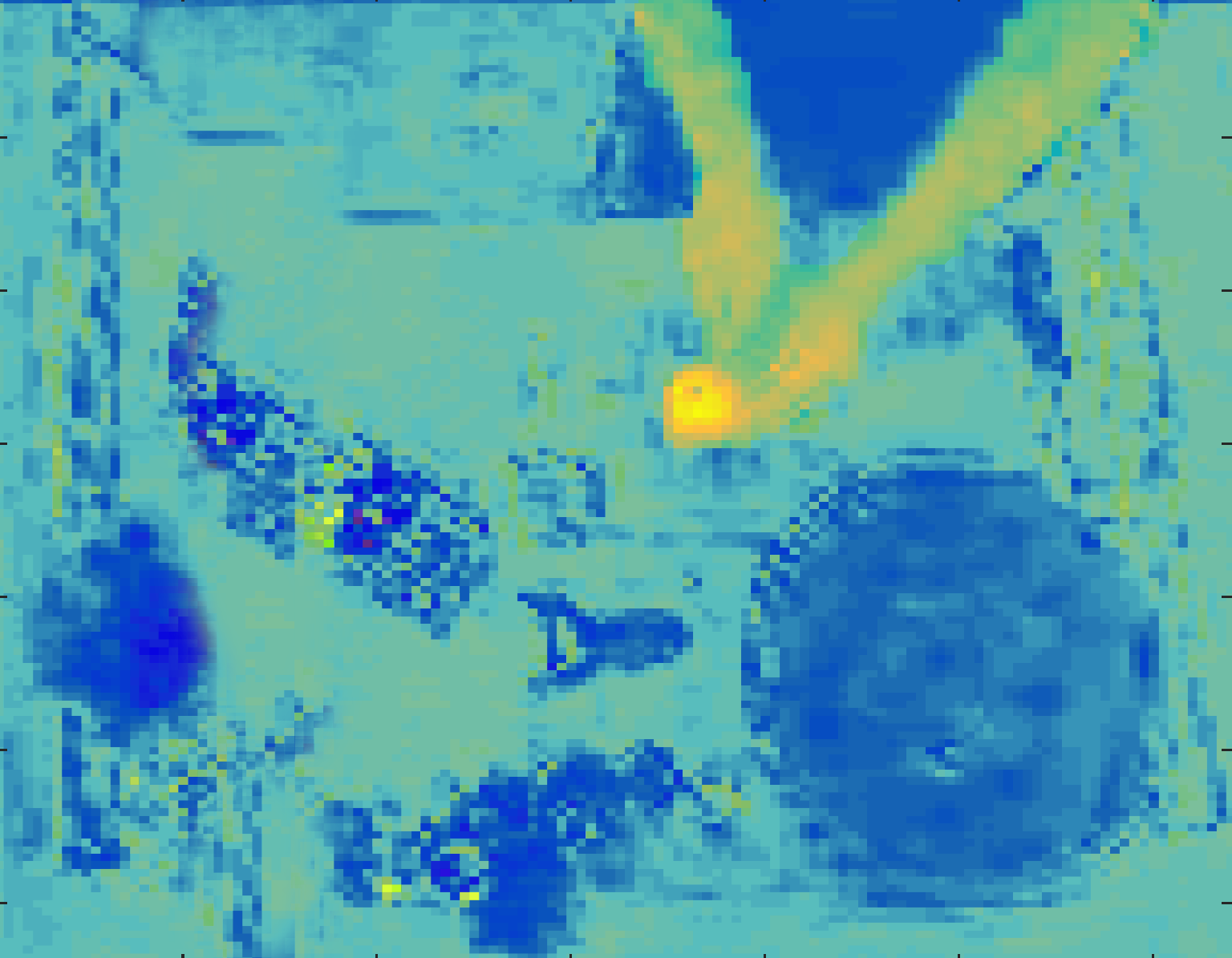}} %
 \end{center}
 \vspace{-7mm}
  \caption{Activation visualisations for the discriminator of the proposed SSA-GAN model.  Yellow denotes more attention while blue denotes less attention.}
  \label{fig:activ_discri}
\end{figure}

With the GAN learning framework, the generator model learns a synthetic objective function that forces it to embed frames from similar action classes close by. This simplifies the task of the action classification process performed by the discriminator model, allowing us to obtain a substantial improvement in performance compared to the baselines. To further demonstrate the discriminator model we obtained activations from the 2nd layer of the discriminator model (see Figure \ref{fig:activ_discri}). In contrast to the generator model, the action classification process of the discriminator is enforced by the generated action codes, hence allowing the discriminator to directly focus on the action specific salient regions even at the early stages such as in layer 2. Hence it is clear that the action classification process is simplified with the proposed GAN approach. 
 
\subsection{Importance of the gated context extractor}

\begin{figure}[!ht]
\begin{center}
    \subfigure[input frame]{\includegraphics[width=.25\textwidth]{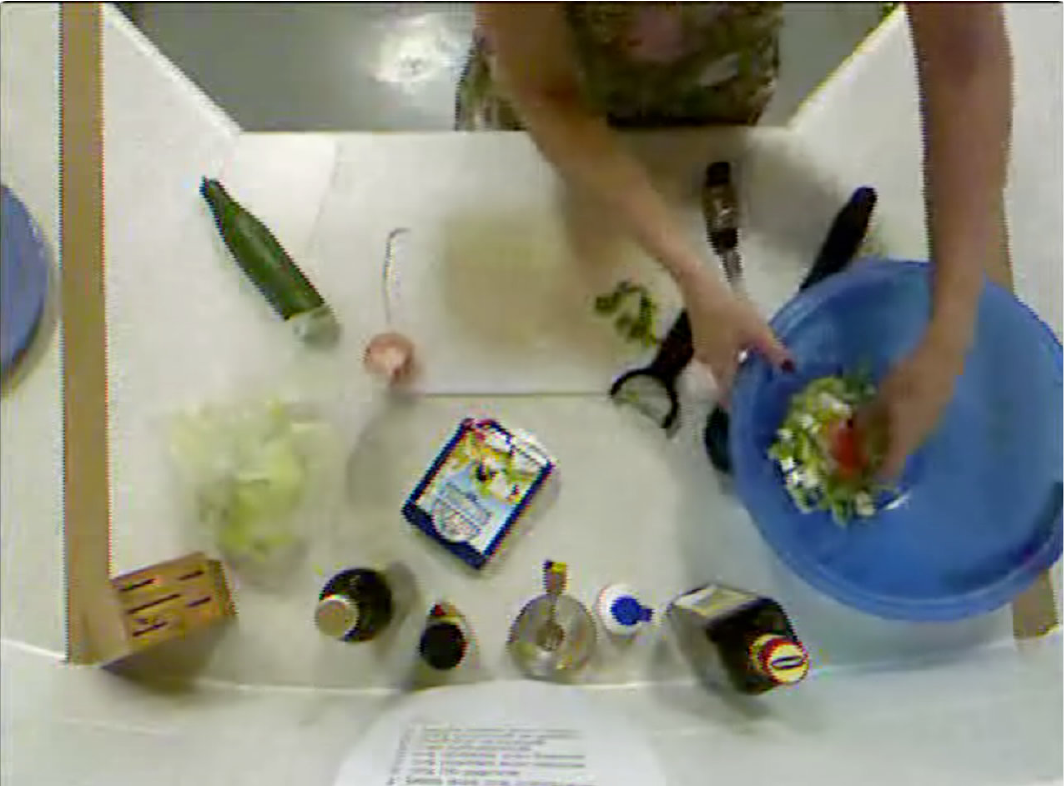}} %
     \subfigure[GCE activations]{\includegraphics[width=.9\textwidth]{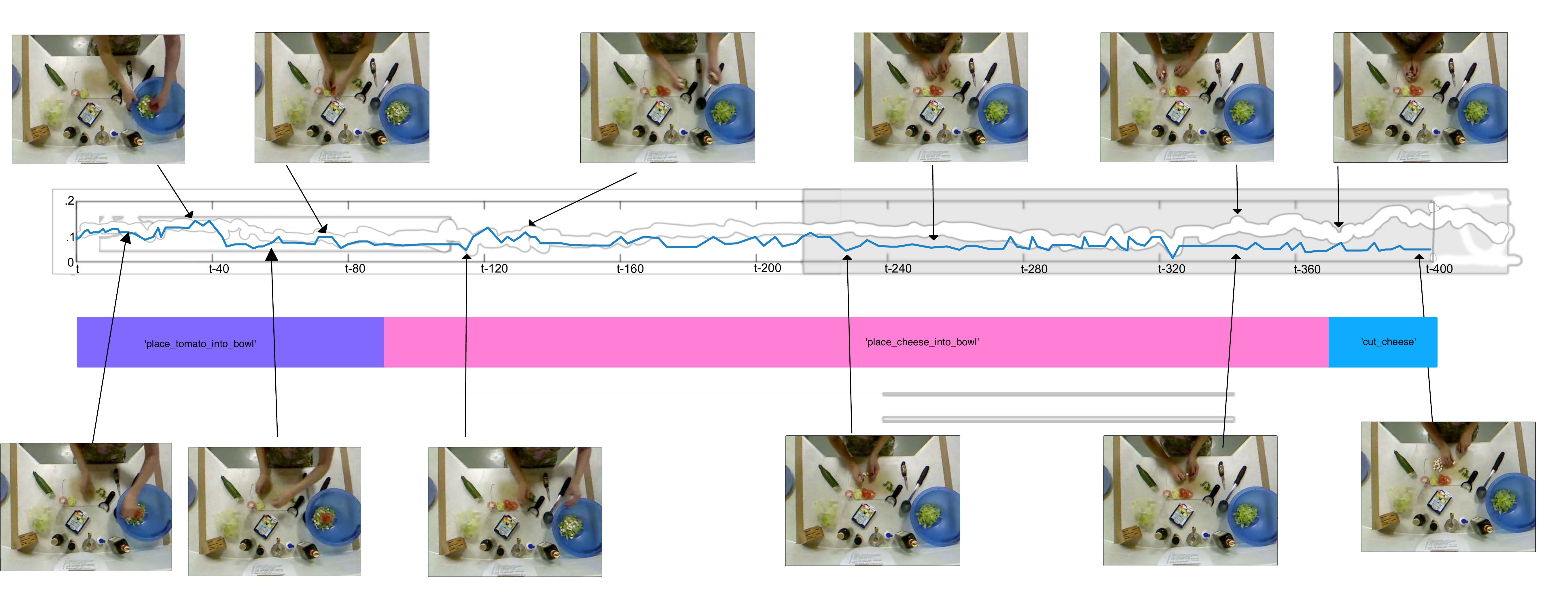}} %
  \end{center}
  \vspace{-7mm}
  \caption{Visualisation of GCE activations for the input frame in (a) for the stored embeddings. The embeddings range from $t$ to t-400 in history. We observe higher activations for embeddings from the short-term and relevant long-term history.}
  \label{fig:GCE_content}
\end{figure}

Figure \ref{fig:GCE_content} shows the distribution of activations from the GCE of the proposed \textit{SSA-GAN} model, for the input frame in Figure \ref{fig:GCE_content} (a). As $m=400$, there exists 400 previous embeddings in the GCE. We denote the current time as t, and thus the GCE content ranges from $t-400$ to $t$. For different peaks and valleys in the activations, we show the embedding that has been stored at the time step.

The GCE module provides higher responses for recent frames as well as for semantically important frame patterns in the long-term history. For instance when recognising the $\mathrm{`place\_tomato\_into\_bowl'} $ action we see higher activations within the short-term history where we see interactions between the actor and the bowl region in the frame, as well as previous interactions in the long-term history where the model has seen hand interactions between the actor and bowl (i.e between $t-240$ to $t-120$), where the actor places cheese into the bowl.

This clearly verifies the importance of efficiently modelling these historical dependencies between the previous actions. The action is not descriptive on its own but it relates to what has happened in the history. Furthermore, the results presented in Table \ref{tab:ablation} further emphasise that it is not sufficient to just to extract out these embeddings from the history, but the module also needs to effectively determine their importance and propagate relevant historic examples to the recognition module (see LSTM and \textit{SSA-GAN}).

\subsection{Time complexity of the proposed \textit{SSA-GAN} model}
We evaluated the computational demands of  the \textit{SSA-GAN} model. The model contains 23M trainable parameters, and outputs 500 predictions in 18.5 seconds using a single core of an Intel E5-2680 2.50 GHz CPU. 

\section{Conclusion}
\label{sec:conclusion}
In this paper, we have proposed a semi-supervised action GAN model (\textit{SSA-GAN}) for fine-grained human action segmentation. The coupled GCE module enables the model to capture the long-term dependencies among previous frames, and exploit the relationships among consecutive action sequences, in order to better model the sub-action level context in the sequence. These innovations enable the model to outperform state-of-the-art methods on three challenging datasets: 50 Salads, MERL Shopping and Georgia Tech Egocentric Activities datasets. The experimental evaluations on video-feeds from cameras with both static-overhead and dynamic-egocentric views, revealed the highly beneficial nature of capturing context information separately, resulting in a significant performance boost and providing the system the flexibility to adapt to the information cues in the different datasets. In addition, extensive evaluations that we have performed on different ablation models demonstrate the importance of the architectural augmentations proposed. It should be noted that even though the model has been evaluated on challenging fine-grained human action datasets, it can also be directly utilised for pre-segmented video action datasets. While the focus of this work is continuous human action segmentation, the application of our proposed \textit{SSA-GAN} is not limited to this. In our future work, we will be investigating the applications of the proposed \textit{SSA-GAN} method for future action prediction which is an important and challenging task for which the proposed technique can be adapted.

\section*{References}

\bibliography{paper4}

\end{document}